\documentclass[twoside,twocolumn,9pt]{article}
\usepackage{extsizes}
\usepackage[super,sort&compress,comma]{natbib}
\usepackage[version=3]{mhchem}
\usepackage[left=1.5cm, right=1.5cm, top=1.785cm, bottom=2.0cm]{geometry}
\usepackage{balance}
\usepackage{mathptmx}
\usepackage{sectsty}
\usepackage{graphicx}
\usepackage{lastpage}
\usepackage[format=plain,justification=justified,singlelinecheck=false,font={stretch=1.125,small,sf},labelfont=bf,labelsep=space]{caption}
\usepackage{float}
\usepackage{fancyhdr}
\usepackage{fnpos}
\usepackage[english]{babel}
\addto{\captionsenglish}{%
  
}
\usepackage{array}
\usepackage{droidsans}
\usepackage{charter}
\usepackage[T1]{fontenc}
\usepackage[usenames,dvipsnames]{xcolor}
\usepackage{setspace}
\usepackage[compact]{titlesec}
\usepackage{hyperref}
\expandafter\def\expandafter\UrlBreaks\expandafter{\UrlBreaks\do\-} 
\usepackage{booktabs}
\usepackage{algorithm}
\usepackage{algpseudocode}
\usepackage{textcomp} 
\usepackage{amsmath} 
\usepackage{amssymb} 
\usepackage{placeins} 

\usepackage{epstopdf}

\definecolor{cream}{RGB}{222,217,201}

\begin{document}

\pagestyle{fancy}
\thispagestyle{plain}
\fancypagestyle{plain}{
\renewcommand{\headrulewidth}{0pt}
}

\makeFNbottom
\makeatletter
\renewcommand\LARGE{\@setfontsize\LARGE{15pt}{17}}
\renewcommand\Large{\@setfontsize\Large{12pt}{14}}
\renewcommand\large{\@setfontsize\large{10pt}{12}}
\renewcommand\footnotesize{\@setfontsize\footnotesize{7pt}{10}}
\makeatother

\renewcommand{\thefootnote}{\fnsymbol{footnote}}
\renewcommand\footnoterule{\vspace*{1pt}%
\color{cream}\hrule width 3.5in height 0.4pt \color{black}\vspace*{5pt}}
\setcounter{secnumdepth}{5}

\makeatletter
\renewcommand\@biblabel[1]{#1}
\renewcommand\@makefntext[1]%
{\noindent\makebox[0pt][r]{\@thefnmark\,}#1}
\makeatother
\renewcommand{\figurename}{\small{Fig.}~}
\sectionfont{\sffamily\Large}
\subsectionfont{\normalsize}
\subsubsectionfont{\bf}
\setstretch{1.125}
\setlength{\skip\footins}{0.8cm}
\setlength{\footnotesep}{0.25cm}
\setlength{\jot}{10pt}
\titlespacing*{\section}{0pt}{4pt}{4pt}
\titlespacing*{\subsection}{0pt}{15pt}{1pt}

\fancyfoot{}
\fancyfoot[LO,RE]{\vspace{-7.1pt}\includegraphics[height=9pt]{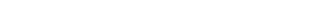}}
\fancyfoot[CO]{\vspace{-7.1pt}\hspace{13.2cm}\includegraphics{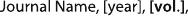}}
\fancyfoot[CE]{\vspace{-7.2pt}\hspace{-14.2cm}\includegraphics{head_foot/RF}}
\fancyfoot[RO]{\footnotesize{\sffamily{1--\pageref{LastPage} ~\textbar  \hspace{2pt}\thepage}}}
\fancyfoot[LE]{\footnotesize{\sffamily{\thepage~\textbar\hspace{3.45cm} 1--\pageref{LastPage}}}}
\fancyhead{}
\renewcommand{\headrulewidth}{0pt}
\renewcommand{\footrulewidth}{0pt}
\setlength{\arrayrulewidth}{1pt}
\setlength{\columnsep}{6.5mm}
\setlength\bibsep{1pt}

\makeatletter
\newlength{\figrulesep}
\setlength{\figrulesep}{0.5\textfloatsep}

\newcommand{\topfigrule}{\vspace*{-1pt}%
\noindent{\color{cream}\rule[-\figrulesep]{\columnwidth}{1.5pt}} }

\newcommand{\botfigrule}{\vspace*{-2pt}%
\noindent{\color{cream}\rule[\figrulesep]{\columnwidth}{1.5pt}} }

\newcommand{\dblfigrule}{\vspace*{-1pt}%
\noindent{\color{cream}\rule[-\figrulesep]{\textwidth}{1.5pt}} }

\makeatother

\twocolumn[
  \begin{@twocolumnfalse}
{\includegraphics[height=30pt]{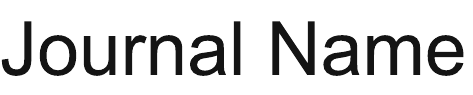}\hfill\raisebox{0pt}[0pt][0pt]{\includegraphics[height=55pt]{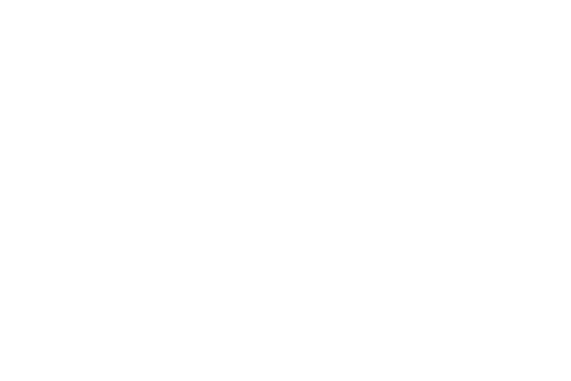}}\\[1ex]
\includegraphics[width=18.5cm]{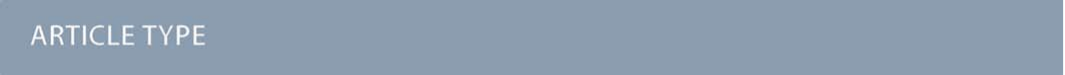}}\par
\vspace{1em}
\sffamily
\begin{tabular}{m{4.5cm} p{13.5cm} }

\includegraphics{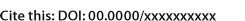} & \noindent\LARGE{\textbf{Low-Cost Autonomous Liquid Handling with Browser-Based Remote Supervision and Closed-Loop Experimental Design (RAINBOT\textsuperscript{TM})$^{\dag}$}} \\
\vspace{0.3cm} & \vspace{0.3cm} \\

 & \noindent\large{Mohamed Rami Ayeche,\textit{$^{a}$} Souhil Sid,\textit{$^{b}$} Ahyen Mostofa,\textit{$^{b}$} Rehaan Hussain,\textit{$^{b}$} Yasmine Abdelkarim,\textit{$^{b}$} Ali Shayesteh$^{\ast}$\textit{$^{a}$} and Fedwa El Mellouhi$^{\ast}$\textit{$^{c}$}} \\

\includegraphics{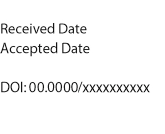} & \noindent\normalsize{Color formulation is a compact but genuinely multivariable test bed for autonomous experimentation and a gateway to the reformulation problems faced in textiles, cosmetics, coatings, plastics, and printing. We present RAINBOT\textsuperscript{TM}, a consumer Cartesian 3D printer converted into a liquid handler that reproduces the deck layout, labware, and operations of a commercial OT-2 for US\$1400. RAINBOT\textsuperscript{TM} is controlled from Python, operated remotely through a browser-based digital twin with a human in the loop, and closes the formulation loop with smartphone-camera color measurement and the CEID\textsuperscript{TM} (Cooperative Explorer for Inverse Design) Bayesian optimization engine. Gravimetric tests over three days ($n = 9$--10 per volume) placed the mean delivered volume within 2\% of nominal from 200 to 1000~\textmu L, with coefficients of variation of 1.1--4.5\%. Mixtures are scored by their CIEDE2000 color difference ($\Delta E_{00}$), the industry-standard color-difference formula defined in ISO/CIE 11664-6, against tolerances derived from replicate preparations of each target. Across three targets and 69 physical evaluations, grass green came closest: the model-guided phase reduced the best $\Delta E_{00}$ from 43.0 after initialization to 1.6 within 24 trials. This was closer to the target than the target recipe itself when it was re-prepared during the campaign (3.6), although still above the empirical tolerance of 1.1. Petroleum blue and pumpkin orange reached 3.5 and 3.3. In a ten-well emerald green replicate study, measured against a designated reference well immediately before the run, the largest distance of a replicate from the reference was 3.6 and the largest distance between any two replicates was 5.1, both inside an operating threshold of 5.5. Repeat imaging of five single-batch colors in the same ten wells gave mean pairwise distances of 2.8--3.6 in the evening, rising to 4.2--5.0 for blue and orange in daylight. Shifts of about 3.5 $\Delta E_{00}$ units for nominally identical recipes between reference preparation and optimization indicate that measurement and preparation stability, rather than the optimizer, limit matching accuracy. We outline the calibration and validation steps needed to approach industrial acceptance thresholds.} \\

\end{tabular}

 \end{@twocolumnfalse} \vspace{0.6cm}
]


\renewcommand*\rmdefault{bch}\normalfont\upshape
\rmfamily
\section*{}
\vspace{-1cm}


\footnotetext{\textit{$^{a}$~Acceleration Consortium, University of Toronto, Toronto, Canada. E-mail: ali.shayesteh@utoronto.ca}}
\footnotetext{\textit{$^{b}$~AISCIA Informatics, Qatar Science and Technology Park, Doha, Qatar}}
\footnotetext{\textit{$^{c}$~Hamad Bin Khalifa University (HBKU), Education City, Doha, Qatar. E-mail: felmellouhi@hbku.edu.qa}}
\footnotetext{\dag~Supplementary Information available: hardware assembly guide, control software, campaign data and camera frames, and a demonstration video. See DOI: 00.0000/00000000.}


\section*{Introduction}

Color formulation provides a compelling model system for demonstrating the transition from conventional trial-and-error materials development toward autonomous, data-efficient experimentation. Achieving a prescribed color is inherently a multivariable problem in which pigment identity and concentration interact nonlinearly, while practical formulations must increasingly satisfy requirements extending well beyond color accuracy. In our self-driving color-mixing platform, the CEID\textsuperscript{TM} (Cooperative Explorer for Inverse Design) Bayesian optimization framework orchestrates an iterative closed loop in which formulations are proposed, physically prepared, characterized, and used to inform subsequent experiments. While target-color matching offers a visually intuitive demonstration of this autonomous workflow,\cite{Baird2022,Lo2024} its broader significance lies in the ability to navigate constrained formulation spaces with limited experimental data. This capability opens opportunities in textiles, cosmetics, paints and coatings, plastics and masterbatches, and printing and packaging, where manufacturers increasingly need to reformulate products in response to sustainability objectives, regulatory restrictions, raw-material availability, cost, and performance requirements; the withdrawal of titanium dioxide (E171) as a food colorant in the European Union is one recent example of a regulatory change that forced reformulation.\cite{EU2022TiO2} Color mixing can therefore serve as a gateway problem for a broader class of self-driving formulation laboratories in which Bayesian optimization is used not merely to reproduce a target property, but to discover alternative formulations that balance performance, environmental impact, manufacturability, and economic constraints.

Accurate and repeatable industrial color matching supports product acceptance, brand consistency, and batch quality in textiles, cosmetics, plastics, automotive coatings, printing, and packaging. Industrial measurement can use calibrated spectrophotometers to characterize spectral response under defined measurement conditions; the Lovibond TR Series is one commercial example.\cite{Lovibond} Formal product acceptance requires an application-specific specification and independent measurement appropriate to the material. The acquisition, calibration, maintenance, and trained operation of such instruments motivate the investigation of more accessible methods for early formulation and screening.

Material color mixing differs from the addition of colored light. In pigment systems, wavelength-dependent absorption and scattering interact with concentration, particle dispersion, opacity, substrate, gloss, and application thickness. A linear combination of precursor RGB values is therefore at best an approximate predictor of a mixture's color, and any such model must be tested against physical mixtures.

The color objective in this work is the CIEDE2000 color difference, the industry-standard formula defined jointly by the International Organization for Standardization (ISO) and the International Commission on Illumination (CIE) in ISO/CIE 11664-6.\cite{ISO116646,Luo2001,Sharma2005} It is computed from CIELAB coordinates $(L^*, a^*, b^*)$, in which $L^*$ is lightness (0 for black to 100 for white), $a^*$ runs from green (negative) to red (positive), and $b^*$ runs from blue (negative) to yellow (positive). For two colors, the standard defines
\begin{equation}
\Delta E_{00} = \sqrt{\left(\frac{\Delta L'}{k_L S_L}\right)^{2} + \left(\frac{\Delta C'}{k_C S_C}\right)^{2} + \left(\frac{\Delta H'}{k_H S_H}\right)^{2} + R_T\,\frac{\Delta C'}{k_C S_C}\,\frac{\Delta H'}{k_H S_H}},
\label{eq:ciede}
\end{equation}
where $\Delta L'$, $\Delta C'$, and $\Delta H'$ are the differences in lightness, chroma, and hue after the CIEDE2000 adjustment of $a^*$; $S_L$, $S_C$, and $S_H$ are weighting functions that correct for the non-uniformity of CIELAB; $R_T$ is a rotation term for the blue region; and $k_L = k_C = k_H = 1$ are the parametric factors for reference conditions.\cite{ISO116646,Sharma2005} $\Delta E_{00}$ is dimensionless, is reported in $\Delta E_{00}$ units, and equals zero only for identical coordinates. The American Association of Textile Chemists and Colorists (AATCC) procedure EP14 recommends CIEDE2000 as the primary method for the instrumental evaluation of small color differences.\cite{AATCCEP14} ASTM D2244-25 addresses instrumental color differences and tolerances for opaque specimens and requires the purchaser and seller to agree on the permissible tolerance and calculation procedure, taking material and appearance conditions into account.\cite{ASTMD2244} Sector-specific procedures exist as well, for example for textiles\cite{ISO105J03} and offset printing.\cite{ISO126472} Acceptance thresholds are therefore application dependent, as the examples in Table~\ref{tbl:thresholds} illustrate.

\begin{table}[t]
\small
\caption{\ Illustrative examples of industrial color acceptance thresholds. Actual tolerances are agreed for each material and measurement condition.\cite{ASTMD2244} The display entry is only an illustrative project comparison, because CIEDE2000 has defined applicability conditions and is not a general-purpose acceptance rule for emissive displays;\cite{ISO116646} cosmetics additionally require attention to gloss, translucency, substrate, and application thickness}
\label{tbl:thresholds}
\begin{tabular*}{0.48\textwidth}{@{\extracolsep{\fill}}ll}
\hline
Application example & Illustrative $\Delta E_{00}$ target \\
\hline
Automotive coatings & $\leq$ 0.5 \\
Cosmetics and skin-tone products & $\leq$ 1.0 \\
Demanding textile matching & $\leq$ 1.5 \\
Packaging and brand colors & $\leq$ 2.0 \\
Offset printing & $\leq$ 2.0 \\
Digital display comparison & $\leq$ 3.0 \\
\hline
\end{tabular*}
\end{table}

This work investigates whether inexpensive RGB measurements combined with Bayesian optimization can reduce the number of physical experiments needed to reverse engineer a target color, and whether this can be done on affordable hardware. The campaigns use liquid food colorants and water rather than industrial pigment dispersions, and their observed color also depends on dilution, liquid depth, background, illumination, and camera response. They therefore provide a model system for a closed-loop formulation proof of concept; they do not establish transfer to coatings, textiles, or cosmetics, where material and appearance effects differ.

Self-driving laboratories (SDLs), which couple automated experimentation with algorithmic decision-making, have shown how quickly closed loops can move through chemical and materials design spaces,\cite{Tom2024,Abolhasani2023,Hase2019} yet their uptake is still limited by the cost of capable liquid handlers and supporting infrastructure.\cite{Lo2024} Much recent work on accessible automation therefore repurposes inexpensive motion hardware: computer-numerical-control machines, consumer 3D printers, and three-axis gantries have been turned into liquid handlers and characterization stations,\cite{Monterrubio2025,Quinn2024} the Jubilee machine offers a reconfigurable open-hardware base,\cite{Vasquez2020} and open designs such as a Python-controlled fraction collector show that research-grade components can be assembled cheaply.\cite{Wang2026} Closed-loop optimization has meanwhile been demonstrated mainly on capable but costly instruments,\cite{Zheng2025} and high-fidelity digital twins of chemistry laboratories have been developed for workflow simulation rather than as lightweight control layers for inexpensive hardware.\cite{Darvish2026} As autonomy grows, the ability of a human expert to observe, trust, and intervene in an experiment becomes more important for reliable and responsible operation.\cite{Amirian2025,Dale2025}

Here we present RAINBOT\textsuperscript{TM} (Real-time Autonomous Intelligent Network for Bidirectional Optimization and Traceability), a consumer Cartesian 3D printer converted into a low-cost liquid handler that reproduces the deck layout, labware, and operations of the Opentrons OT-2.\cite{OpentronsOT2} Converting a printer into a liquid handler has precedents; our contribution is an OT-2-equivalent platform costing US\$1400 (Table~\ref{tbl:bom}) that integrates (i) Python-controlled liquid handling, (ii) a browser-based digital twin through which the robot is monitored and operated remotely with a human in the loop, and (iii) closed-loop experimental design in which the CEID\textsuperscript{TM} Bayesian optimization engine learns from smartphone-camera color measurements. We applied the system to three target colors and report in detail the campaign that came closest, grass green; the other two are documented in the Appendix.

\section*{Problem formulation}

The research problem is to identify a reproducible formulation within an application-specific color tolerance, using a limited number of physical preparations, when the measurement system itself introduces variability. Search efficiency and reliable acceptance must therefore be evaluated together.

Let $x = (R, Y, B, W)$ denote a recipe, where $R$, $Y$, $B$, and $W$ are the integer numbers of 100~\textmu L units of red, yellow, and blue colorant and water. The experimental domain $\mathcal{X}$ is the set of all such recipes in which each component $x_i$ lies between 0 and 5 units and the four components sum to 10 units:
\begin{equation}
\mathcal{X} = \{x \in \mathbb{Z}^4 : 0 \leq x_i \leq 5,\; R + Y + B + W = 10\}.
\label{eq:domain}
\end{equation}
Every mixture therefore has a volume of 1000~\textmu L. Without the upper bound there are $\binom{13}{3} = 286$ ways to split 10 units among four components; removing the $4\binom{7}{3} = 140$ splits in which one component exceeds 5 units leaves 146 feasible recipes. Let $F(x; c)$ denote the true color of recipe $x$, expressed as CIELAB coordinates $(L^*, a^*, b^*)$, when it is prepared and measured under the conditions $c$ (lighting, camera position, plate, and settling time), and let $\ell_\mathrm{target}$ denote the CIELAB coordinates of the reference (target) color. The desired formulation is the recipe whose color difference from the target, eqn~(\ref{eq:ciede}), is smallest:
\begin{equation}
x^* \in \arg\min_{x \in \mathcal{X}} \Delta E_{00}\bigl(F(x; c),\, \ell_\mathrm{target}\bigr).
\label{eq:objective}
\end{equation}
Operationally, the aim is to find a recipe whose color difference is at or below a tolerance $T$ within a fixed experimental budget, and then to test that candidate using independent preparations. The optimizer does not observe $F(x; c)$ directly. It observes a camera estimate: each trial's score is obtained from the componentwise median of ten RGB vectors, converted to CIELAB, and compared with the saved target.

Before the physical campaigns, the workflow was tested on the digital twin, using Bayesian optimization on a synthetic nonlinear mixing function. Bayesian optimization converged within 10--15 iterations, which indicated that a physical budget of 24 experiments, 16.4\% of the search space, was a reasonable allocation. The physical targets were set to grass green, petroleum blue, and pumpkin orange, offering enough diversity to explore the proportions of the precursors and the darkness of the mixtures.

\section*{Experimental}

\subsection*{Platform overview}
\begin{figure*}[t]
\centering
\includegraphics[width=\textwidth]{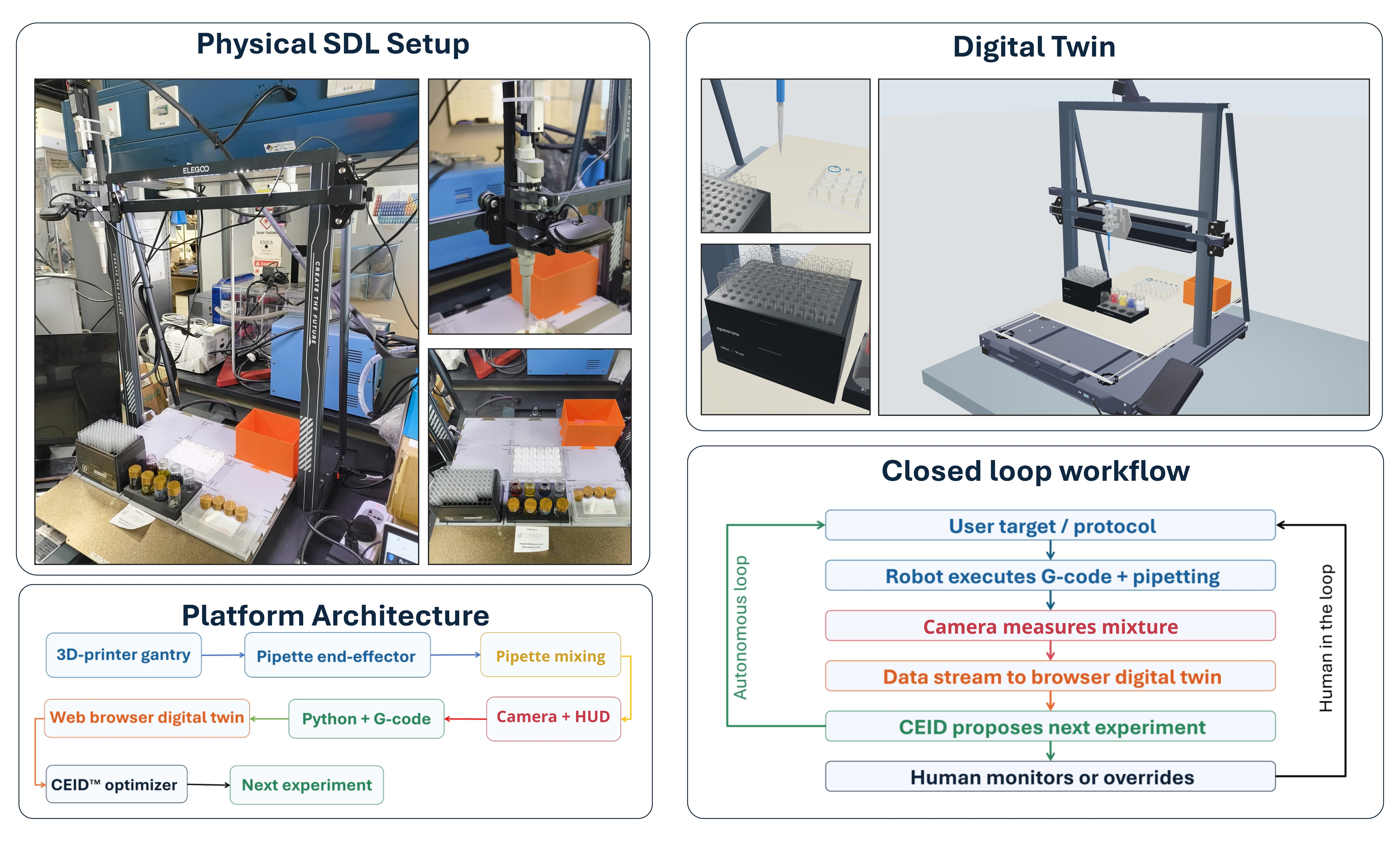}
\caption{Overview of RAINBOT\textsuperscript{TM}: the converted 3D-printer platform, its browser-based digital twin, the platform architecture, and the closed-loop workflow with a human in the loop. A Python and G-code layer drives the gantry and pipette, the camera head-up display (HUD) measures each mixture, and CEID\textsuperscript{TM} Bayesian optimization proposes the next experiment while the operator monitors or overrides the system. The HUD is the live camera overlay that shows the well sampling circles, RGB and CIEDE2000 readouts, gantry coordinates, and well-concentricity indicators on top of the camera feed.}
\label{fig:overview}
\end{figure*}

RAINBOT\textsuperscript{TM} has four parts: a modified 3D-printer gantry that provides programmable X--Y--Z motion; a single-channel pipette end-effector with motorized plunger and tip-eject actuation; a smartphone camera that measures the color of each well; and a software stack that orchestrates motion and measurement, mirrors the platform in a browser-based digital twin, and hosts the CEID\textsuperscript{TM} decision loop. Fig.~\ref{fig:overview} shows the assembled system and the closed-loop workflow.

\subsection*{Hardware}
\subsubsection*{Motion platform.~~}
RAINBOT\textsuperscript{TM} is built on an Elegoo Neptune 4 Max fused-deposition 3D printer, a consumer Cartesian machine with a 420~$\times$~420~$\times$~480~mm working volume and pre-installed Klipper firmware. Its open architecture allows the mechanics to be modified freely. We removed the extruder assembly and mounted a pipette carriage in its place, so the gantry's X--Y--Z motion positions the pipette tip directly over the deck, which is laid out with the slot positions and labware of an OT-2.

\subsubsection*{Pipette end-effector.~~}
A precision Eppendorf Research plus single-channel pipette (100--1000~\textmu L) is mounted on the carriage. Two compact linear actuators (15~mm\,s$^{-1}$ travel speed, 60~N force) operate it mechanically: a 30~mm-stroke actuator presses the plunger (aspirate/dispense) button, and a 50~mm-stroke actuator presses the tip-eject button. Tips are picked up by lowering the pipette onto the tip rack. To aspirate, the plunger button is pressed, the tip is lowered into the source liquid, and the button is released to draw liquid up; to dispense, the button is pressed again. Repeating these aspirate--dispense cycles in the destination well mixes its contents. Because the actuators drive the pipette's own plunger against its mechanical stops, the delivered volume inherits the pipette's intrinsic tolerance (see Results and discussion).

\subsubsection*{Actuation and imaging.~~}
The two linear actuators are switched through an eight-channel USB relay board (SainSmart) driven from the host computer over a serial connection; the spare channels leave room for further actuators or instruments. Hook-up wiring and a few 3D-printed adapters complete the assembly. Color is measured by a smartphone camera mounted above the 24-well plate, which streams 1280~$\times$~720 video to the control computer through DroidCam with optical zoom, focus, and lighting fixed on the phone. The bill of materials is given in Table~\ref{tbl:bom}, and the assembly guide is provided in the Supplementary Information.\dag

\begin{table}[!b]
\small
\caption{\ Bill of materials for RAINBOT\textsuperscript{TM}. Costs are approximate prices in US dollars at the time of writing, and the product source for each item is cited}
\label{tbl:bom}
\begin{tabular*}{0.48\textwidth}{@{\extracolsep{\fill}}lll}
\hline
Component & Description & Cost / USD \\
\hline
Elegoo Neptune 4 Max\cite{ElegooN4Max} & Cartesian gantry, Klipper & 560 \\
Eppendorf Research plus\cite{EppendorfRPlus} & 100--1000 \textmu L pipette & 575 \\
Xiaomi Redmi smartphone\cite{RedmiPhone} & Camera, DroidCam stream & 130 \\
Linear actuator (30 mm)\textsuperscript{\textit{a}} & Plunger drive & 20 \\
Linear actuator (50 mm)\textsuperscript{\textit{a}} & Tip-eject drive & 20 \\
SainSmart relay board\cite{SainSmartRelay} & 8-channel USB & 30 \\
3D-printed parts\textsuperscript{\textit{b}}\cite{ElegooPLA} & Deck plates, racks, mounts & 45 \\
Wiring and fasteners & Hook-up wire, screws & 20 \\
\hline
Total & & 1400 \\
\hline
\multicolumn{3}{l}{\footnotesize\textsuperscript{\textit{a}} Generic 12~V micro linear actuators (15~mm\,s$^{-1}$, 60~N).}\\
\multicolumn{3}{l}{\footnotesize\textsuperscript{\textit{b}} About 3~kg of PLA filament at US\$13.99 per kg, plus print wastage.}\\
\end{tabular*}
\end{table}

\subsection*{Equivalence with the OT-2}
RAINBOT\textsuperscript{TM} was designed as a functional equivalent of the OT-2 (Table~\ref{tbl:equivalence}). It uses the OT-2's slot numbering and positions and the same labware definitions, performs the same pipetting operations over the same volume range, and receives the same commands from the experiment agent, which sends a recipe and receives a measurement regardless of which robot executes it. The main practical differences are the motion layer, G-code on Klipper rather than the Opentrons robot server, and speed: RAINBOT\textsuperscript{TM} runs at roughly a quarter of the OT-2's pace.

\begin{table}[t]
\small
\caption{\ Functional comparison of RAINBOT\textsuperscript{TM} with the Opentrons OT-2}
\label{tbl:equivalence}
\begin{tabular*}{0.48\textwidth}{@{\extracolsep{\fill}}lll}
\hline
Feature & OT-2 & RAINBOT\textsuperscript{TM} \\
\hline
Base & Commercial handler & Converted 3D printer \\
Deck layout & OT-2 slots & Same slots \\
Labware & Opentrons definitions & Same definitions \\
Pipette range & 100--1000 \textmu L & 100--1000 \textmu L \\
Operations & Tip, aspirate, dispense, & Same \\
 & mix, eject & \\
Control & Python, robot server & Python, G-code \\
Agent interface & Recipe in, color out & Same \\
Relative speed & 1 & $\approx$0.25 \\
Cost / USD\textsuperscript{\textit{a}} & $\approx$20\,000\textsuperscript{\textit{b}} & $\approx$1400 \\
\hline
\end{tabular*}
\par\vspace{2pt}\noindent\begin{minipage}{0.48\textwidth}\footnotesize\textsuperscript{\textit{a}} For comparison, the Opentrons Flex\cite{OpentronsOT2} costs at least US\$26\,400, and commercial liquid handlers such as Tecan and Hamilton systems cost at least US\$50\,000. \textsuperscript{\textit{b}} OT-2 list price with one pipette fitted (US\$18\,900--23\,850 depending on the pipettes; base robot US\$15\,950).\cite{OpentronsOT2}\end{minipage}
\end{table}

\subsection*{Software}
\subsubsection*{Motion orchestration.~~}
A custom Python library coordinates each experiment. Gantry motion is commanded by streaming G-code to the printer's Klipper firmware, and pipetting is performed by timed activation of the relay channels. We modified the Klipper configuration to home the Z-axis sensorlessly to the top of the printer, which keeps the pipette clear of labware during setup and travel.

\subsubsection*{Browser-based digital twin.~~}
\begin{figure}[t]
\centering
\includegraphics[width=\columnwidth]{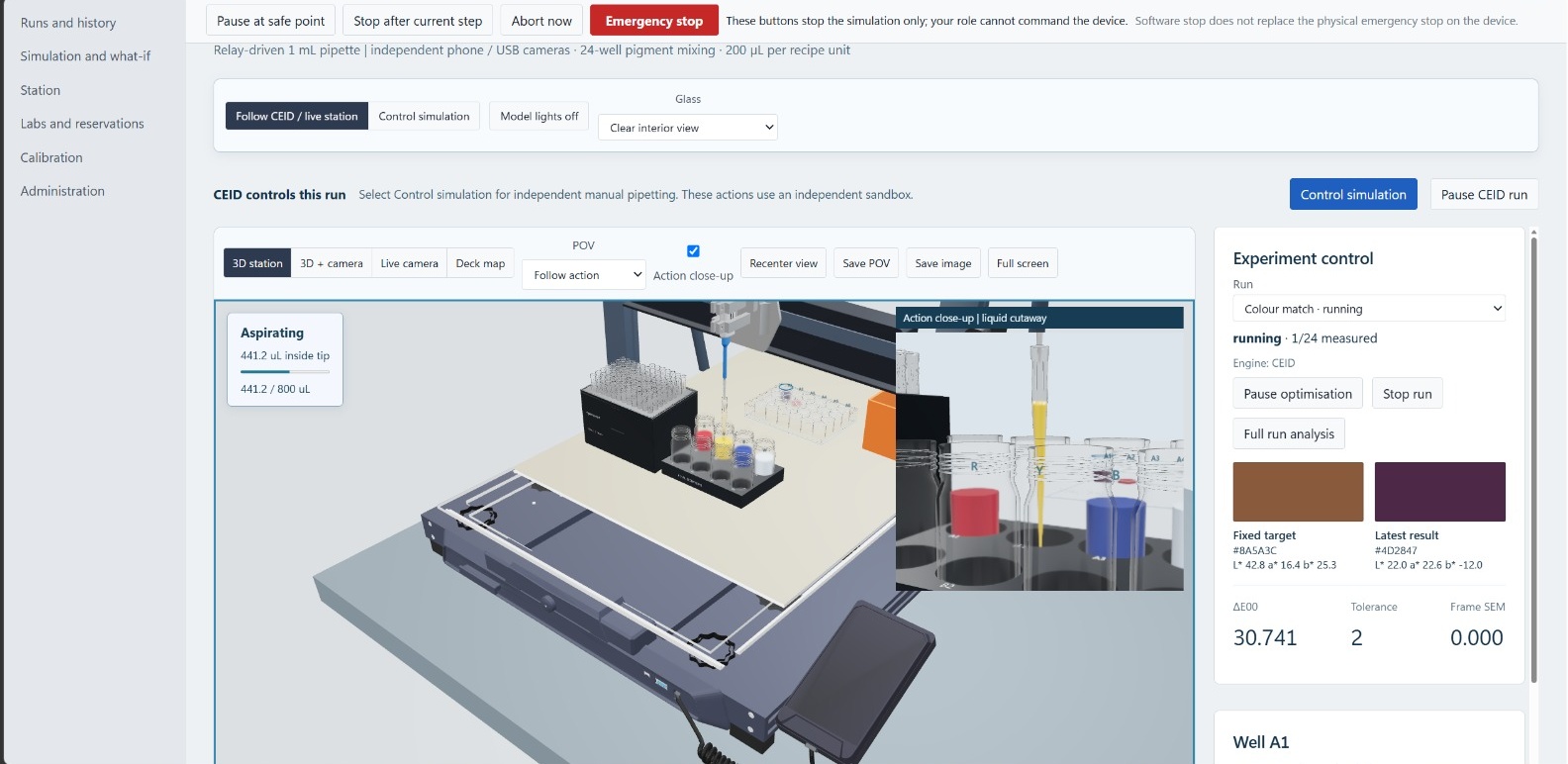}
\caption{Control panel of the browser-based digital twin of RAINBOT\textsuperscript{TM} during a CEID\textsuperscript{TM}-controlled run. The 3D station view mirrors the gantry, pipette, and deck in real time; the close-up shows the liquid drawn into the tip during aspiration; and the experiment-control panel shows the fixed target, the latest result, its $\Delta E_{00}$, and the tolerance. The `Emergency stop' button is the remote software stop described in the text.}
\label{fig:twin}
\end{figure}

The digital twin is a WebGL web application that runs in any modern browser without installation (Fig.~\ref{fig:twin}). The Python layer that drives RAINBOT\textsuperscript{TM} also updates the twin: each G-code command sent to the gantry is mirrored as motion of the virtual 3D model, and pipetting events (aspirate, dispense, eject) and live measurements are streamed bidirectionally over a WebSocket connection. The twin is also the remote control surface: an operator can start, continue, or stop a run, issue manual commands, and follow each measurement from anywhere. In its present form the twin mirrors the commanded state of the robot; an on-board Logitech C270 webcam records the true position of the carriage alongside the commanded G-code coordinates, so that physical--virtual agreement can be checked independently.

The human remains in the loop throughout. The operator defines targets and tolerances, monitors every measurement, and can intervene at any time; in the initial camera run on RAINBOT\textsuperscript{TM}, each trial proposed by CEID\textsuperscript{TM} also had to be approved by the operator before it was executed.

\subsubsection*{Camera head-up display (HUD) and color pipeline.~~}
\begin{figure*}[t]
\centering
\includegraphics[width=\textwidth]{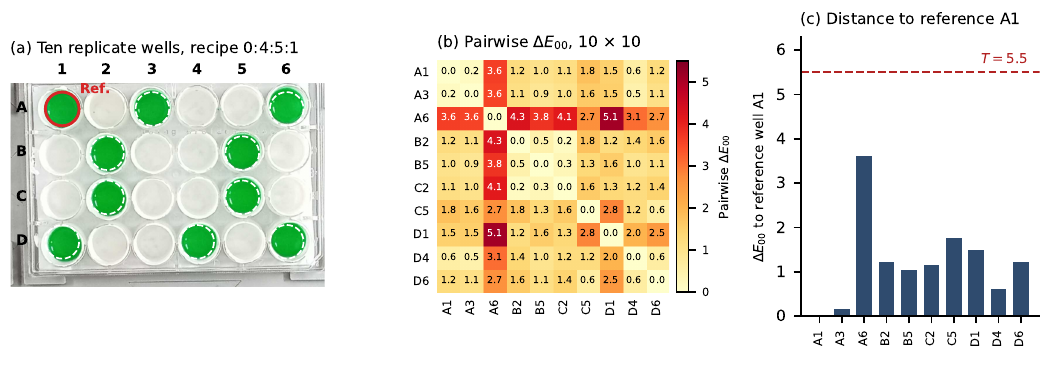}
\caption{Ten-well emerald green measurement check with reference well A1 (mobile phone camera). (a) Full-plate image at target verification: ten wells hold independent preparations of the emerald green recipe 0:4:5:1, and well A1 (red ring) is the designated reference; the remaining wells hold water. (b) Symmetric $10 \times 10$ matrix of pairwise $\Delta E_{00}$ between the ten replicate wells. (c) $\Delta E_{00}$ of each replicate to the reference well A1, with the operating threshold $T = 5.5$ (dashed line). Values are rounded to one decimal place; full-precision values are given in Appendix~F.}
\label{fig:hud}
\end{figure*}

The HUD is the live camera overlay that displays the well sampling circles, the RGB and CIEDE2000 readouts, the gantry coordinates, and well-concentricity indicators on top of the camera feed. After the robot has cleared the view and a five-second settling delay has elapsed, the HUD reads the phone stream and collects ten fresh RGB vectors from a circular sampling region fitted to each well of the 24-well plate (Fig.~\ref{fig:hud}a). Their componentwise median defines the observed color. Each 8-bit channel is divided by 255 and interpreted in the standard RGB (sRGB) color space of IEC 61966-2-1, the default encoding of consumer cameras.\cite{IEC61966} The values are then converted to CIE XYZ tristimulus values, the device-independent coordinates defined by the CIE, and then to CIELAB. Both conversions use the CIE standard illuminant D65, which represents average daylight, and the CIE 1931 2$^\circ$ standard observer, which describes average human color matching over a 2$^\circ$ field of view. Finally, the difference from the fixed target is computed with eqn~(\ref{eq:ciede}) using the reference parametric factors $k_L = k_C = k_H = 1$.\cite{ISO116646,Luo2001,Sharma2005} These steps use the open-source Colour Science for Python library. The score is the $\Delta E_{00}$ of the median RGB rather than the mean of the ten per-frame scores. As a check, we recomputed every target, tolerance, noise level, and per-trial score reported here from the logged RGB values with an independent implementation validated against the published CIEDE2000 test data;\cite{Sharma2005} all agreed to within 0.01 $\Delta E_{00}$ units. Camera-derived $L^*a^*b^*$ values are sRGB-based estimates without spectrophotometric traceability.

\subsubsection*{Data logging.~~}
Motion commands, pipetting states, camera frames, recipes, and scores are time-stamped and logged, both for live display in the digital twin and for archiving and offline analysis. A local agent reserves each trial's identifier, destination well, and tips in an SQLite journal before any robot motion, so that an interrupted run can be reconciled afterward. If the optimizer repeats a request for a trial that has already been completed, the agent returns the saved measurement instead of dispensing again.

\subsection*{Closed-loop optimization with CEID\textsuperscript{TM}}
CEID\textsuperscript{TM} sends integer recipes and trial identifiers to the experiment agent over a local transmission control protocol (TCP) network connection; the agent validates each recipe, reserves wells and tips, and issues the robot commands. Each campaign has a single objective: to drive the CIEDE2000 difference between the measured mixture and the target, $\Delta E_{00}$, toward zero. With $e$ the measured $\Delta E_{00}$ of a trial and $T$ the campaign tolerance, both in $\Delta E_{00}$ units, CEID\textsuperscript{TM} works with the normalized distance
\begin{equation}
D = \frac{\max(0,\, e - T)}{\max(1,\, e,\, T)}.
\label{eq:distance}
\end{equation}
Acceptance requires $e \leq T$, which gives $D = 0$ and triggers a stop. Because every tolerance used here is at least 1, any trial above tolerance also has $e > 1$, and eqn~(\ref{eq:distance}) reduces to $D = 1 - T/e$, which lies between 0 and 1. $D$ is dimensionless and increases monotonically with $e$, so we report the measured $\Delta E_{00}$ as the interpretable measure of mismatch.

The optimizer searches the finite set $\mathcal{X}$ of eqn~(\ref{eq:domain}), excludes recipes that have already been measured, and scales each component to the unit interval by dividing by five. The first $N_0 = 6$ physical evaluations are drawn from a Sobol sequence, a low-discrepancy sequence that spreads points evenly over the domain.\cite{Sobol1967} A Gaussian-process (GP) surrogate model (BoTorch SingleTaskGP) is then fitted to all available observations with the supplied observation noise,\cite{Balandat2020} and the next batch of $q$ distinct recipes is selected with the batch noisy log expected improvement acquisition function, qLogNEI.\cite{Ament2023} Candidates within a batch are proposed jointly and measured serially, and the model is updated between batches (Algorithm~\ref{alg:ceid}). In the grass green campaign the Sobol points were scrambled with seed 42 and projected onto the nearest unmeasured feasible recipe, and the GP modeled the negative normalized objective with standardized outputs, a noise variance propagated from the reference measurements and floored at $10^{-6}$, at most 75 fitting iterations, and 128 quasi-Monte Carlo (QMC) acquisition samples; these detailed settings are recorded for grass green and are not assumed to be identical for the other campaigns.

\begin{algorithm}[t]
\caption{CEID\textsuperscript{TM} closed-loop color optimization}
\label{alg:ceid}
\begin{algorithmic}[1]
\Require Target color $\ell_\mathrm{target}$; feasible recipe set $\mathcal{X}$; number of Sobol points $N_0 = 6$; total budget $N$; batch size $q$; tolerance $T$
\State Draw $N_0$ Sobol points and project them onto $\mathcal{X}$ to give $X_0$.
\State Set experimental dataset $\mathcal{D}$ to empty.
\For{each $x$ in $X_0$}
  \State Prepare $x$; measure its median camera color $\ell(x)$.
  \State $e \gets \Delta E_{00}(\ell(x),\, \ell_\mathrm{target})$; append $(x, e)$ to $\mathcal{D}$.
  \If{$e \leq T$} \Return $x$ \EndIf
\EndFor
\While{$|\mathcal{D}| < N$}
  \State Fit the Gaussian-process surrogate to $\mathcal{D}$.
  \State Select $q$ unmeasured recipes in $\mathcal{X}$ by qLogNEI.
  \For{each selected $x$}
    \State Prepare $x$; measure $\ell(x)$; $e \gets \Delta E_{00}(\ell(x),\, \ell_\mathrm{target})$.
    \State Append $(x, e)$ to $\mathcal{D}$.
    \If{$e \leq T$} \Return $x$ \EndIf
  \EndFor
\EndWhile
\State \Return $x_{\text{best}} = \arg\min_{\mathcal{D}} e$,\quad $e_{\text{best}} = \min_{\mathcal{D}} e$,\quad $\mathcal{D}$.
\end{algorithmic}
\end{algorithm}

\subsection*{Campaign protocols}
\subsubsection*{Composition and preparation.~~}
Recipes use R:Y:B:W for red, yellow, and blue food colorant and water, in 100~\textmu L units summing to 10 (eqn~(\ref{eq:domain})). The reservoir holds red, yellow, blue, and water in positions A1--A4, and mixtures are prepared in a 24-well plate whose wells are labeled by rows A--D and columns 1--6. Each nonzero component is transferred with a fresh tip, and the tip of the last component mixes the well with four 700~\textmu L aspirate--dispense cycles before disposal.

\subsubsection*{Target definition and tolerance.~~}
For grass green, the target recipe 0:5:2:3 (500~\textmu L yellow, 200~\textmu L blue, 300~\textmu L water) was prepared independently in ten wells, and each well was read ten times. The fixed target is the componentwise median of the ten per-well median RGB vectors, RGB [89.5, 169.5, 11] ($L^* = 62.3$, $a^* = -47.6$, $b^* = 62.1$). The tolerance is the linearly interpolated empirical 95th percentile of the ten well-to-target distances, $T = 1.09$, which nine of the ten wells satisfy. Because the same wells define both the target and the tolerance, $T$ is an in-sample operating threshold, not validated 95\% future coverage. Across the 45 distinct pairs of reference wells, the median $\Delta E_{00}$ was 0.9 (range 0.3--2.0); preparation and spatial effects were not separated. The sample standard deviation (SD) of the ten reference scores, 0.3, was supplied to the optimizer as a shared observation-noise level. Petroleum blue and pumpkin orange followed the same construction with four and five reference wells; for both, a production tolerance of 2.0 was selected (see Appendix). The target must be measured immediately before each production run, because the reference color drifts between occasions (see Results and discussion).

\subsubsection*{Initial camera run on RAINBOT\textsuperscript{TM}.~~}
The camera loop was first run on RAINBOT\textsuperscript{TM} against a green target captured as RGB [28, 154, 17]. Each component ranged from 0 to 5 units with the sum limited to 10, and the optimizer minimized $\Delta E_{00}$ directly with six Sobol trials, one recipe per batch, and a random seed of 42; the operator approved each trial before it was executed. This run predates the replicate-based target and tolerance protocol.

\subsubsection*{Execution of the production campaigns.~~}
Because RAINBOT\textsuperscript{TM} runs at roughly a quarter of the OT-2's speed, the liquid handling for the three production campaigns was carried out on an OT-2 to complete them within two days, using the same deck layout, labware, recipes, mixing protocol, camera, HUD, agent, and CEID\textsuperscript{TM} configuration that RAINBOT\textsuperscript{TM} executes (Table~\ref{tbl:equivalence}). The results transfer to RAINBOT\textsuperscript{TM} for three reasons. First, with recipes quantized in 100~\textmu L units, the dispensing hardware affects the measured color only through the delivered volume of each component and the completeness of mixing. Second, RAINBOT\textsuperscript{TM} delivers 200--1000~\textmu L with mean errors within 2\% and coefficients of variation (CV) of 1.1--4.5\% (Table~\ref{tbl:dispense}), so a component volume deviates by about 0.1 recipe units at one standard deviation. Near the best grass green recipe, exchanging one unit between two components shifts $\Delta E_{00}$ by 0.5 to 7.3 (Table~\ref{tbl:greentrials}), so this volume error corresponds to roughly 0.05--0.7 $\Delta E_{00}$. This is below the median spread between replicate reference preparations (0.9) but not always below the noise level supplied to the optimizer (0.3). Third, RAINBOT\textsuperscript{TM} performs the same four-cycle mixing with its 100--1000~\textmu L pipette, and the complete camera loop has been run on RAINBOT\textsuperscript{TM} itself (initial camera run). Repeating the calibrated campaigns directly on RAINBOT\textsuperscript{TM} is nonetheless included among the validation priorities.

\subsubsection*{Ten-well emerald green measurement check.~~}
To test whether the camera measurement can be trusted on RAINBOT\textsuperscript{TM} itself, the emerald green recipe 0:4:5:1 (400~\textmu L yellow, 500~\textmu L blue, 100~\textmu L water) was prepared independently in ten wells (A1, A3, A6, B2, B5, C2, C5, D1, D4, and D6), and the other 14 wells received water only (Fig.~\ref{fig:hud}a). The plate was imaged with the mobile phone camera immediately before the run, and each well was read ten times. Well A1 was designated as the reference well, and every replicate was compared with it and with every other replicate. All 24 wells were then read in six batches of four through the same agent and scoring pipeline used in the optimization campaigns, against the pooled target RGB [8.2, 170.3, 40.4] (the componentwise median of the ten replicates) and an operating threshold of $T = 5.5$. This threshold was set to exceed the largest pairwise distance between replicate wells (Results and discussion).

\subsubsection*{Multi-color ten-well camera repeatability.~~}
To separate camera and plate-position effects from preparation, five colors were each prepared by hand as a single batch from their R:Y:B:W ratios (grass green 0:5:2:3, petroleum blue 1:0:5:4, pumpkin orange 2:5:0:3, orange-red 3:5:0:2, and emerald green 0:4:5:1) and dispensed at 1000~\textmu L into the same ten wells as the emerald check; the robot did not handle liquid. Each plate was imaged with the mobile phone camera (flash off, fluorescent white balance, exposure +9, manual focus) from the detection position, and each well was read ten times. Every color was measured in the evening with the printer's built-in lights off and on, and petroleum blue, pumpkin orange, and grass green were repeated the next morning in daylight from an unshaded window. Well A1 was the reference. A color passed overall if the mean of its 45 pairwise $\Delta E_{00}$ values was below $T = 5.5$, and an individual well failed if its distance to A1 was 5.5 or more.

\section*{Results and discussion}

\subsection*{A low-cost platform}
The converted printer works as a programmable liquid handler while keeping the low cost of its consumer base. The pipette carriage reuses the gantry's repeatable X--Y--Z motion, and driving the pipette's own plunger with external actuators avoids any bespoke fluidics. The complete RAINBOT\textsuperscript{TM} hardware, including the smartphone camera, costs US\$1400 (Table~\ref{tbl:bom}), most of it for the printer and the research-grade pipette. A refurbished pipette and a discounted printer bring a functional build to roughly US\$1000 without changing the architecture. That is less than a tenth of the price of an OT-2 configured with a single-channel pipette, whose functions it reproduces, and far below the Opentrons Flex and commercial automation platforms (Table~\ref{tbl:equivalence}). Replacing a spectrophotometer with a smartphone camera keeps the measurement side equally accessible. The control and digital-twin software are available on request (see Data availability); the CEID\textsuperscript{TM} engine and the specific hardware implementation are proprietary to AISCIA Informatics, so the platform is best described as low-cost and built from commodity components rather than as fully open hardware.

\subsection*{Programmable, accurate dispensing}
\begin{figure}[t]
\centering
\includegraphics[width=\columnwidth]{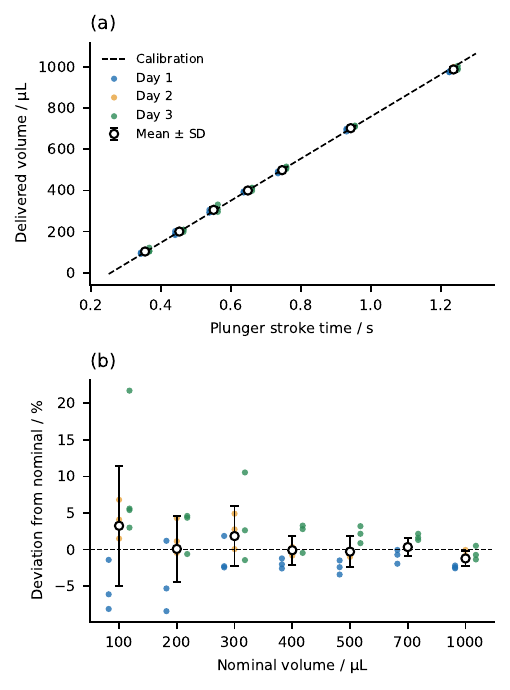}
\caption{Gravimetric validation of RAINBOT\textsuperscript{TM} at seven nominal volumes over three days (evaporation-corrected; $n = 9$--10 per volume). (a) Delivered volume against plunger stroke time; the dashed line is the calibration. (b) Deviation from nominal volume. Colored points are individual dispenses by day, offset horizontally for clarity, and open circles are means $\pm$1 sample SD.}
\label{fig:gravimetric}
\end{figure}

Because the actuators drive the plunger against its native stops, RAINBOT\textsuperscript{TM} inherits the metrological behavior of the pipette rather than adding a fluidic path of its own. Volume is set by the actuation distance, which is controlled by the duration of relay activation at the actuator's fixed speed. The relay timing was calibrated gravimetrically: the plunger stroke time is 0.256~s plus 0.098~s per 100~\textmu L (0.354~s at 100~\textmu L and 1.236~s at 1000~\textmu L), and the aspirate and dispense pulses are 0.10~s longer than the stroke. A previous timing over-dispensed by 8--59\% across 100--1000~\textmu L and was replaced by this calibration. Validation used deionized water dispensed into a 24-well plate on an OHAUS Explorer analytical balance (220~g capacity, 0.1~mg readability, internal AutoCal calibration). The balance was not tared. Each dispensed mass was the difference between consecutive cumulative readings, converted to volume with a density of 0.998~mg\,\textmu L$^{-1}$ and corrected for evaporation from the plate between readings at a rate measured on each day (1.6--1.8~mg\,min$^{-1}$). Seven nominal volumes from 100 to 1000~\textmu L were tested in mixed order on three days, with three dispenses per volume per day and a fourth at 100~\textmu L on the third day (Table~\ref{tbl:dispense}, Fig.~\ref{fig:gravimetric}; individual values in Table~\ref{tbl:rawbalance}). From 200 to 1000~\textmu L the mean delivered volume was within 2\% of nominal, with CVs of 1.1--4.5\% that fall as the volume increases. At 100~\textmu L the mean was 3.2\% high with a CV of 7.9\%. Day means differed by a few percent, with the first day low and the third day high at most volumes. Four first attempts on the second day were repeated and replaced: one produced no dispense, one was short, and two used washed tips that still held drops of water. A residual drop of about 5--20~\textmu L changes a 100~\textmu L dispense by 5--20\% but a 1000~\textmu L dispense by only 1--2\%, so tips must be dry before small-volume transfers. The evaporation correction assumes a constant rate and slightly over-corrects early readings, so uncorrected means are also given.

\begin{table}[t]
\footnotesize
\caption{\ Plunger stroke time and gravimetric validation over three days (mean $\pm$ sample standard deviation of the evaporation-corrected volume, with the uncorrected mean; water density 0.998~mg\,\textmu L$^{-1}$)}
\label{tbl:dispense}
\begin{tabular*}{0.48\textwidth}{@{\extracolsep{\fill}}rrrrrrr}
\hline
Target & Stroke & $n$ & Volume & Error & CV & Uncorrected \\
/ \textmu L & / s & & / \textmu L & / \% & / \% & / \textmu L \\
\hline
100  & 0.354 & 10 & $103.2 \pm 8.2$  & $+3.2$ & 7.9 & 96.8 \\
200  & 0.452 & 9  & $200.2 \pm 9.0$  & $+0.1$ & 4.5 & 194.3 \\
300  & 0.550 & 9  & $305.5 \pm 12.3$ & $+1.8$ & 4.0 & 298.7 \\
400  & 0.648 & 9  & $399.7 \pm 7.9$  & $-0.1$ & 2.0 & 392.2 \\
500  & 0.746 & 9  & $498.6 \pm 10.6$ & $-0.3$ & 2.1 & 492.6 \\
700  & 0.942 & 9  & $702.4 \pm 8.7$  & $+0.3$ & 1.2 & 696.2 \\
1000 & 1.236 & 9  & $987.9 \pm 10.6$ & $-1.2$ & 1.1 & 978.8 \\
\hline
\end{tabular*}
\end{table}

\subsection*{Remote operation through the digital twin}
The digital twin mirrored RAINBOT\textsuperscript{TM} throughout operation. Comparing the webcam record of the carriage with the displayed positions showed sub-second latency between a completed physical motion and its appearance in the twin, so an expert could follow and steer each run remotely. When the twin issued a stop signal, the Python motion controller halted execution before the next queued motion; we refer to this as a remote software stop.

\subsection*{Closed-loop color formulation}
\subsubsection*{Initial camera run on RAINBOT\textsuperscript{TM}.~~}
In the first camera-in-the-loop run on RAINBOT\textsuperscript{TM}, the third Sobol trial, 0:4:5:0, came within $\Delta E_{00} = 2.4$ of the green target in a run of 24 trials, and none of the remaining 21 trials came closer (Fig.~\ref{fig:camera}a; Table~\ref{tbl:run2}). The first model-guided proposals stayed near this recipe, changing one or two components at a time, before moving to more distant regions. The run showed that the full loop of robot, camera, digital twin, and optimizer operates on RAINBOT\textsuperscript{TM}. It also showed that, without a replicate-based target and tolerance, there was no way to tell how much of the variation between nearby recipes was real. That observation led to the calibrated protocol used in the production campaigns.

\subsubsection*{Grass green.~~}
\begin{figure}[t]
\centering
\includegraphics[width=\columnwidth]{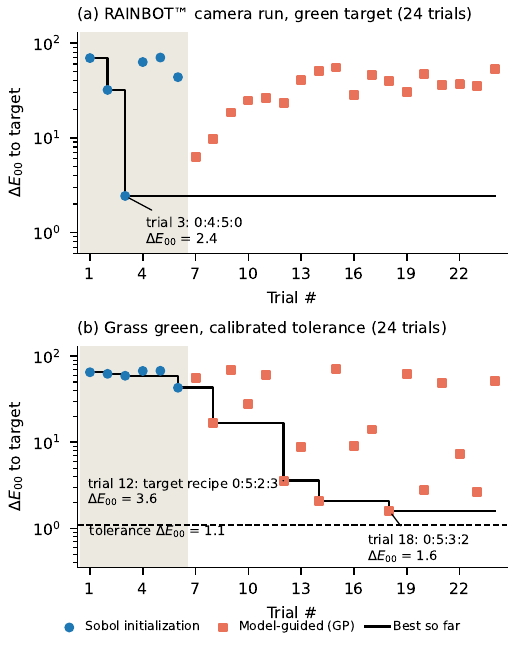}
\caption{Closed-loop color formulation with CEID\textsuperscript{TM}. (a) Initial camera run on RAINBOT\textsuperscript{TM} against a green target, before the replicate-based tolerance protocol. (b) Grass green production campaign with the tolerance calibrated from ten replicate preparations (dashed line, $\Delta E_{00} = 1.1$). Shaded regions mark the six Sobol initialization trials, squares the model-guided (GP) trials, and the black line the best value so far; error bars in (b) are bootstrap 2.5--97.5 percentile ranges and are mostly smaller than the markers.}
\label{fig:camera}
\end{figure}

\begin{figure}[t]
\centering
\includegraphics[width=\columnwidth]{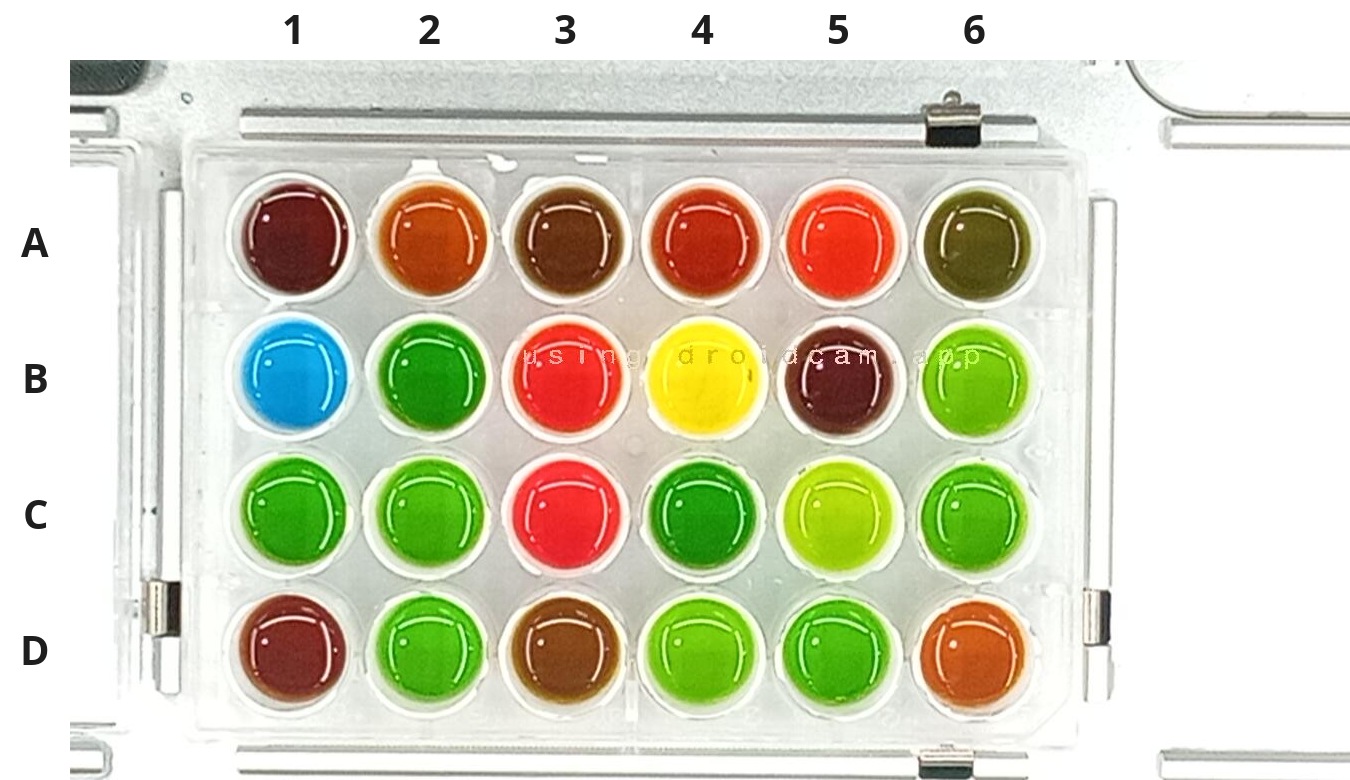}
\caption{Grass green plate after trial 24, with rows A--D and columns 1--6 labeled. Trials 1--24 fill A1 to D6 in row order. C6 holds the best mixture (trial 18, 0:5:3:2), and B6 holds the exact target recipe (trial 12, 0:5:2:3). The photograph retains acquisition reflections and the DroidCam watermark.}
\label{fig:plategreen}
\end{figure}

The accepted grass green campaign contained 24 unique recipes in eight batches of three: six Sobol and 18 model-guided trials (Fig.~\ref{fig:camera}b, Fig.~\ref{fig:plategreen}, Table~\ref{tbl:milestones}). All six Sobol recipes contained red and were far from the target. The first model-guided batch tested two-component recipes on the edges of the domain and reached 16.7 with 0:5:5:0, and the best score then fell from 43.0 after initialization to 1.6 at trial 18 in well C6 (0:5:3:2), a 96.3\% reduction. Five of the 18 model-guided trials came within $\Delta E_{00} = 5$ of the target, against none of the six Sobol trials, and all five contained no red. These stage statistics come from one run in which the model-guided phase also had the Sobol data available, so they do not by themselves establish that the optimizer outperforms an alternative; that requires repeated runs with matched budgets. The 24 recipes covered 16.4\% of the domain.

\begin{table}[t]
\footnotesize
\caption{\ Grass green milestones (complete results in Table~\ref{tbl:greentrials})}
\label{tbl:milestones}
\begin{tabular*}{0.48\textwidth}{@{\extracolsep{\fill}}lllp{2.6cm}}
\hline
Milestone & Trial \#, well & Best $\Delta E_{00}$ & Recipe or note \\
\hline
End of Sobol & 6, A6 & 43.0 & Initial recipes far from target \\
First value below 20 & 8, B2 & 16.7 & 0:5:5:0 \\
Exact target recipe & 12, B6 & 3.6 & 0:5:2:3; did not reproduce the target RGB \\
Further improvement & 14, C2 & 2.1 & 0:4:3:3 \\
Final best & 18, C6 & 1.6 & 0:5:3:2 \\
Budget exhausted & 24, D6 & 1.6 & Final two batches did not improve \\
\hline
\end{tabular*}
\end{table}

No trial met the tolerance. The best result is 0.5 above $T$, or about 1.5 times the threshold, and its temporal bootstrap range, 1.5--1.6, lies entirely above it. It nonetheless sits in the target's color region: its lightness ($L^* = 62.0$ versus 62.3) and chroma (78.9 versus 78.2) are almost identical to the target's, and it differs mainly in hue, by 2.8$^\circ$ toward blue-green. Against the illustrative thresholds of Table~\ref{tbl:thresholds}, a difference of 1.6 lies within the examples for packaging, offset printing, and display comparison and outside those for demanding textiles, cosmetics, and automotive coatings. This is only a numerical comparison; camera-derived values cannot establish compliance with any industrial specification.

The exact target recipe 0:5:2:3 was evaluated at trial 12 in well B6 and gave RGB [109, 179, 19] and $\Delta E_{00} = 3.6$, whereas the same well position had read [89, 169, 13] during target calibration. The optimizer's best recipe, with one more unit of blue and one fewer of water, therefore matched the target more closely than the target recipe did under production conditions. By optimizing the measured appearance, CEID\textsuperscript{TM} compensated for a change in the physical or optical response between calibration and production, although the data do not isolate preparation, elapsed-time, or optical effects. At about three times the tolerance, this shift is the clearest measure of what currently limits matching accuracy.

The campaign included two documented recoveries. After trial 9, an interrupted preparation in well B4 was discarded and the run resumed from the nine saved observations with fresh tips. At trial 18 the C6 mixture was prepared, but camera acceptance failed after the sampling circle shifted; the liquid was retained, the circle re-centered, and ten fresh readings replaced the failed capture without dispensing again. The accepted dataset contains exactly one result per trial, with no repeated initialization.

\subsubsection*{Measurement uncertainty.~~}
The campaigns distinguish three observed variation measures from the surrogate's predictive uncertainty. The frame SD, the sample standard deviation of the ten per-frame scores, describes short-term camera variation for an existing mixture and ranged from 0 to 0.4 across the grass green trials. The temporal bootstrap resamples the ten RGB vectors as whole triplets, recomputes the median-based score, and gave SDs of at most 0.3 (50\,000 resamples); its 2.5--97.5 percentile ranges are conditional on a fixed target and are not validated confidence limits. The noise level supplied to the optimizer, 0.3 for grass green, is the standard deviation of the reference scores and is not the posterior standard deviation of the Gaussian process. Per-candidate posterior and acquisition values were not retained. The bootstrap assumes exchangeable readings and excludes target estimation error, calibration bias, drift, preparation variability, and spatial effects, and quantized RGB values can yield zero spread without implying zero physical uncertainty. A full uncertainty budget would follow established practice for spectrophotometric results,\cite{ASTME2867} using independent repeat preparations, reference or control samples, and a calibrated instrument.

\subsection*{Three target colors}
Table~\ref{tbl:colors} summarizes the three campaigns; the complete results for petroleum blue and pumpkin orange are given in the Appendix. Grass green and petroleum blue improved their best scores substantially within their runs, by 96.3\% and 85.1\% from the end of initialization. Pumpkin orange reached a cluster of similar final colors, but changes between its online and final measurement occasions of up to 8.8 $\Delta E_{00}$ units limit inference about the optimizer's improvement, and its best Sobol and model-guided results differ by less than 0.01. None met its campaign threshold. Petroleum blue's best trial was its own reference recipe, 1:0:5:4, re-prepared during the campaign, and it measured 3.5 from the target. Together with the 3.6 of the re-prepared grass green recipe, this points to a recurring shift of about 3.5 $\Delta E_{00}$ between reference preparation and optimization, larger than either tolerance. Recipe search can compensate for part of such a shift, as grass green shows, but it cannot replace a stable measurement. Cross-color scores cannot be used as a controlled ranking of optimization performance, because targets, tolerance construction, deck and camera arrangement, batches, and measurement occasions differ, and one run per color cannot establish success rates, repeated-seed performance, or a global optimum.

\begin{table}[t]
\footnotesize
\caption{\ Summary of the three production campaigns. Recipes are R:Y:B:W; $T$ is the campaign tolerance and the bootstrap range is the temporal 2.5--97.5 percentile range of the best score}
\label{tbl:colors}
\begin{tabular*}{0.48\textwidth}{@{\extracolsep{\fill}}llll}
\hline
 & Grass green & Petroleum blue & Pumpkin orange \\
\hline
Date (2026) & 28 Oct & 29 Oct & 29 Oct \\
Reference recipe & 0:5:2:3 & 1:0:5:4 & 2:5:0:3 \\
Reference wells & 10 & 4 & 5 \\
$T$ & 1.1 & 2.0 & 2.0 \\
Trials (Sobol/GP) & 24 (6/18) & 24 (6/18) & 21 (6/15) \\
Best trial \#, well & 18, C6 & 24, B6 & 20, D2 \\
Best recipe & 0:5:3:2 & 1:0:5:4 & 4:3:0:3 \\
Best $\Delta E_{00}$ & 1.6 & 3.5 & 3.3 \\
Best $D$ & 0.32 & 0.44 & 0.39 \\
Bootstrap range & 1.5--1.6 & 3.1--3.6 & 3.3--3.3 \\
\hline
\end{tabular*}
\end{table}

\subsection*{Ten-well emerald green measurement check}
The ten-well check measured how far independent preparations of one recipe spread under the camera when the target is measured immediately before the run (Fig.~\ref{fig:hud}; complete data in Appendix~F). The largest distance of a replicate from the reference well A1 was $\Delta E_{00} = 3.6$ (well A6), and the other eight replicates lay within 1.8 of A1. Across the 45 pairs of replicate wells, the median distance was 1.3 (range 0.2--5.1), and every pair was within the operating threshold of 5.5. The largest distances all involved well A6, a corner well whose mixture read lighter and more saturated than the others ($L^* = 65.0$ and chroma $C^* = 83.0$, against 60.8 and 79.5 for A1); without A6 the largest pairwise distance was 2.8. The 14 water wells scored 33.2--35.7 against the pooled target, at least six times the threshold, so the threshold separated every replicate ($D = 0$) from every blank ($D = 0.83$--0.85). The frame SD did not exceed 0.15 in any of the 24 wells, so short-term camera noise was small compared with the spread between wells.

The spread between independent preparations, up to about 5 $\Delta E_{00}$ units within a single plate, is comparable to the shift of about 3.5 units observed for re-prepared reference recipes in the optimization campaigns. With the target measured immediately before the run and a threshold of 5.5, the camera measurement reproduced the replicate recipe in every well and rejected every blank, which supports its use in production runs. The threshold of 5.5 is an operating criterion for this set-up; it is much wider than the industrial examples of Table~\ref{tbl:thresholds} and does not by itself establish matching accuracy at those tolerances.

\subsection*{Multi-color ten-well camera repeatability}
With one batch per color in the same ten wells, every evening run passed overall, with mean pairwise $\Delta E_{00}$ of 2.8--3.6 with the lights on or off (Table~\ref{tbl:multicolor}). Individual pairs reached 6.4--7.9, so the maximum exceeded 5.5 in every run, and 34--42 of the 45 pairs were below the threshold. Wells failing against A1 recurred at the same positions, B2, B5, C5, and D6, across colors, while the frame SD stayed at or below 0.24. Because the liquid was identical in every well, this spread reflects plate position and illumination rather than preparation. It is at least as large as the spread between the independent emerald preparations (median 1.3, maximum 5.1), which were imaged on a different day. In daylight, petroleum blue and pumpkin orange rose to mean pairwise distances of 4.2--5.0 with maxima of 9.6--13.7, and up to nine wells failed against A1, while grass green with the lights on gave the best run (mean 2.1, maximum 5.5). Camera measurements should therefore be made with the plate shaded from daylight or in the evening, and plate position should be treated as a source of variation.

\section*{Limitations and perspectives}
Connecting the physical experiment and the digital twin supports remote experimentation with human oversight. The following limitations and research directions should guide further development and discussion with the community.
\begin{itemize}
\item \textbf{Action space and experimental control.} The current recipe space is bounded and discrete. Future work could expand the twin's controllable parameters to include mixing, settling time, and acquisition settings, while distinguishing recipe variables from environmental and measurement controls.
\item \textbf{Interfaces and observations.} A documented Python API could support remote interaction with the physical robot and its digital twin, exposing experimental controls, observations, calibration status, and uncertainty estimates. The interface should make clear how physical observations update the twin and how proposed actions are checked before execution.
\item \textbf{Autonomous calibration.} Calibration before each production run could assess measurement stability and identify unsuitable conditions before optimization begins. Its effectiveness requires independent validation, including tests across plate positions and measurement occasions, and its results could inform whether a run proceeds or requires operator review.
\item \textbf{Surrogate models.} Surrogates could reduce the number of physical experiments by guiding candidate selection and updating predictions from measured outcomes. Their benefit should be evaluated against defined baselines, accounting for model error and measurement uncertainty; the current campaigns do not establish an expected reduction in experimental cost.
\item \textbf{Remote operation and environmental conditions.} Operation across day and night introduces changes in daylight and artificial lighting, and people's shadows, reflections, and movement around the apparatus can also affect camera readings. In the multi-color check, daylight raised the mean pairwise $\Delta E_{00}$ of identical wells from 2.9--3.3 to 4.2--5.0 for petroleum blue and pumpkin orange. Pre-run calibration and continuous monitoring of the surroundings could help detect these changes and reduce uncertainty, but their effectiveness remains to be established. An open question for the community is how to represent environmental conditions within the twin and decide when to correct measurements, recalibrate, or pause an experiment.
\item \textbf{Human oversight.} Human involvement remains valuable for defining objectives and tolerances, reviewing calibration, and resolving unexpected conditions. Future work should clarify which decisions can be automated and when operator intervention is required, particularly during remote operation.
\end{itemize}

Three validation steps have priority. First, the target should be measured immediately before each production run, as in the ten-well check, and a stable physical reference should be repeated at the beginning and end of each run under fixed lighting, exposure, white balance, camera position, and sampling geometry, and independent reference and best-recipe preparations should be randomized across plate positions to distinguish preparation variability, spatial effects, and drift; the recurring outlying positions B2, B5, C5, and D6 make this randomization necessary. Second, acceptance thresholds should be validated using independent wells and later measurement occasions; pumpkin orange's reference variability in particular warrants resolving the measurement process before a two-unit criterion is interpreted as a repeatability bound. Third, camera estimates should be compared with independently measured standards or a calibrated instrument, after which repeated preparations and repeated-seed campaigns, run directly on RAINBOT\textsuperscript{TM}, can support meaningful comparisons of convergence. Simulator transfer should be assessed on a held-out recipe panel before simulated history is used as prior information in a physical campaign. Direct measurements of positioning accuracy, tip-handling reliability, and carryover on RAINBOT\textsuperscript{TM} will complete its characterization.

\section*{Conclusions}
The significance of a self-driving color-mixing laboratory ultimately extends beyond automating the search for a desired shade. Coupling robotic or automated formulation and characterization with CEID\textsuperscript{TM}-driven Bayesian optimization provides a general strategy for transforming color matching into a multi-objective materials-design problem. Future implementations could begin with an established formulation and autonomously identify alternatives that preserve a target color while replacing pigments that are environmentally undesirable, increasingly regulated, costly, supply-constrained, or incompatible with emerging manufacturing requirements. In textiles, this could support lower-impact coloration while retaining fastness; in cosmetics, color equivalence could be pursued alongside safety, stability, sensory, and regulatory requirements; in paints and coatings, weatherability, corrosion protection, and durability could accompany color targets; while plastics, masterbatches, and printing inks introduce additional constraints associated with processing, recyclability, migration, and substrate compatibility. Incorporating spectral measurements, life-cycle indicators, cost, regulatory information, and application-specific performance into the optimization loop would thus move autonomous color formulation from single-objective matching toward the inverse design of sustainable formulations. More broadly, such systems illustrate how accessible, rapidly measurable formulation problems can become proving grounds for self-driving laboratories capable of learning from sparse experimental data and navigating increasingly complex materials design spaces.

\section*{Author contributions}
M.~R.~Ayeche, S.~Sid, A.~Mostofa, and R.~Hussain designed and built the hardware, developed the control software and digital twin, and performed the experiments. A.~Mostofa developed the camera HUD used for color measurement, and S.~Sid carried out the CEID\textsuperscript{TM} campaigns and their analysis. The CEID\textsuperscript{TM} framework was developed at AISCIA Informatics. A.~Shayesteh and F.~El~Mellouhi conceived and supervised the project. All authors contributed to the analysis and to writing and reviewing the manuscript.

\section*{Conflicts of interest}
The CEID\textsuperscript{TM} (Cooperative Explorer for Inverse Design) framework is a proprietary technology of AISCIA Informatics, with which several of the authors are affiliated; the authors declare this commercial interest. The authors declare that there are no patents or patent applications related to the work reported in this manuscript.

\section*{Data availability}
The digital twin software, the per-trial campaign data, and the camera frames are available from the corresponding authors upon reasonable request. The CEID\textsuperscript{TM} optimization engine and the physical hardware implementation are proprietary technologies of AISCIA Informatics and are not distributed under an open-source or open-hardware license.

A reviewer-accessible build of the digital twin can be provided as a browser-based WebGL application. No installation is required; reviewers can open it in a WebGL-compatible desktop browser. The demonstration includes the twin interface and a representative closed-loop workflow but does not provide unrestricted access to the proprietary CEID\textsuperscript{TM} service or the physical robotic hardware. Requests for access may be directed to \href{mailto:info@aiscia.com}{info@aiscia.com}.

\section*{Acknowledgements}
This research was funded in part by the Canada First Research Excellence Fund (Grant No.\ CFREF-2022-00042) awarded to the University of Toronto's Acceleration Consortium. The research was also funded in part by the IBDA Grant 2025 awarded to AISCIA Informatics LLC, a company incubated at the Qatar Science and Technology Park, and by Hamad Bin Khalifa University, Office of Innovation and Industrial Relations, Doha, Qatar.


\balance

\bibliography{rsc}
\bibliographystyle{rsc}

\clearpage
\onecolumn
\setcounter{table}{0}\renewcommand{\thetable}{A\arabic{table}}
\setcounter{figure}{0}\renewcommand{\thefigure}{A\arabic{figure}}
\section*{Appendix}
\noindent All $\Delta E_{00}$ values are CIEDE2000 differences from the fixed campaign target, computed from the componentwise median of ten camera RGB readings. $D$ is the normalized distance of eqn~(\ref{eq:distance}). Frame SD is the sample standard deviation of the ten per-frame $\Delta E_{00}$ scores, and the bootstrap range gives the 2.5th and 97.5th percentiles of the median-based score under resampling of the ten RGB vectors (50\,000 resamples for grass green and 30\,000 for petroleum blue and pumpkin orange). These ranges describe short-term temporal variation only; they exclude target uncertainty, independent preparation variability, calibration bias, and between-occasion drift, and a zero width does not imply zero physical uncertainty. Values are rounded for display: $\Delta E_{00}$, bootstrap ranges, and $L^*a^*b^*$ coordinates to one decimal place, and $D$, frame SD, and reference-well distances to two decimal places; the full-precision records are available with the campaign data. Recipes are R:Y:B:W in 100~\textmu L units, and Sobol and GP denote the initialization and model-guided phases.

\subsection*{A. Grass green}
Table~\ref{tbl:stocks} gives the camera colors of the three stocks, Table~\ref{tbl:greenref} the ten reference preparations that define the grass green target and tolerance, and Table~\ref{tbl:greentrials} all 24 accepted trials.
\begin{table}[!htbp]
\small
\caption{\ Camera colors of the three stocks, each imaged as a separate 1000~\textmu L sample with ten readings. Water was used only as the diluent and was not separately characterized. These values document appearance under the camera arrangement; they are neither absorption spectra nor a physical mixing law}
\label{tbl:stocks}
\begin{tabular*}{0.75\textwidth}{@{\extracolsep{\fill}}lllll}
\hline
Stock & Well & Median camera RGB & HEX & Estimated $L^*a^*b^*$ \\
\hline
Red & A1 & 180, 19, 24 & \#B41318 & 38.3, 59.6, 42.4 \\
Yellow & A2 & 230, 194, 17 & \#E6C211 & 79.3, $-$1.8, 78.9 \\
Blue & A3 & 17, 72, 137.5 & \#11488A & 30.8, 8.8, $-$41.3 \\
\hline
\end{tabular*}
\end{table}

\begin{table}[!htbp]
\small
\caption{\ Grass green reference preparations of 0:5:2:3 (500~\textmu L yellow, 200~\textmu L blue, 300~\textmu L water). The pooled target, RGB [89.5, 169.5, 11], is the componentwise median of the ten well medians; the tolerance $T = 1.09$ is the linearly interpolated 95th percentile of the ten distances}
\label{tbl:greenref}
\begin{tabular*}{0.75\textwidth}{@{\extracolsep{\fill}}lll}
\hline
Well & Median camera RGB & $\Delta E_{00}$ to pooled target \\
\hline
B1 & 85, 168, 11 & 0.71 \\
C1 & 88, 170, 12 & 0.26 \\
D1 & 90, 172, 11 & 0.73 \\
B6 & 89, 169, 13 & 0.19 \\
C6 & 95, 170, 16 & 0.67 \\
D6 & 89, 166, 10 & 1.04 \\
B4 & 86, 166, 12 & 1.13 \\
D4 & 94, 169, 9 & 0.65 \\
B3 & 92, 171, 11 & 0.54 \\
D3 & 93, 172, 10 & 0.86 \\
\hline
\end{tabular*}
\end{table}

\begin{table}[!htbp]
\footnotesize
\caption{\ Grass green: all 24 accepted trials (target RGB [89.5, 169.5, 11], $T = 1.09$, noise level supplied to the optimizer 0.30)}
\label{tbl:greentrials}
\begin{tabular*}{\textwidth}{@{\extracolsep{\fill}}lllllllll}
\hline
Trial \# & Well & Phase & R:Y:B:W & Median camera RGB & $\Delta E_{00}$ & $D$ & Frame SD & Bootstrap 2.5--97.5\% \\
\hline
1 & A1 & Sobol & 4:0:4:2 & 56, 15, 15.5 & 65.1 & 0.98 & 0.10 & 65.0--65.2 \\
2 & A2 & Sobol & 2:4:1:3 & 149.5, 46, 16 & 62.4 & 0.98 & 0.04 & 62.3--62.4 \\
3 & A3 & Sobol & 2:2:4:2 & 73, 32, 22 & 59.2 & 0.98 & 0.08 & 59.2--59.3 \\
4 & A4 & Sobol & 3:2:1:4 & 148, 28, 18 & 67.4 & 0.98 & 0.12 & 67.2--67.4 \\
5 & A5 & Sobol & 3:2:0:5 & 217, 40, 25 & 67.7 & 0.98 & 0.03 & 67.6--67.7 \\
6 & A6 & Sobol & 1:3:5:1 & 66, 55, 25 & 43.0 & 0.98 & 0.31 & 42.5--43.0 \\
7 & B1 & GP & 0:0:5:5 & 18, 131, 187 & 56.0 & 0.98 & 0.00 & 56.0--56.0 \\
8 & B2 & GP & 0:5:5:0 & 23, 126, 16 & 16.7 & 0.94 & 0.01 & 16.7--16.7 \\
9 & B3 & GP & 5:5:0:0 & 220.5, 41, 39 & 69.1 & 0.98 & 0.10 & 69.1--69.2 \\
10 & B4 & GP & 0:5:0:5 & 243, 216, 17 & 27.8 & 0.96 & 0.00 & 27.8--27.8 \\
11 & B5 & GP & 5:0:5:0 & 91, 42, 41 & 61.1 & 0.98 & 0.36 & 61.1--61.9 \\
12 & B6 & GP & 0:5:2:3 & 109, 179, 19 & 3.6 & 0.70 & 0.14 & 3.5--3.8 \\
13 & C1 & GP & 0:5:4:1 & 38, 148, 18 & 8.9 & 0.88 & 0.12 & 8.8--9.0 \\
14 & C2 & GP & 0:4:3:3 & 73, 170, 19 & 2.1 & 0.47 & 0.06 & 2.1--2.1 \\
15 & C3 & GP & 5:0:0:5 & 231, 43, 51 & 70.6 & 0.99 & 0.07 & 70.6--70.6 \\
16 & C4 & GP & 0:4:5:1 & 38, 148, 24 & 9.0 & 0.88 & 0.09 & 9.0--9.0 \\
17 & C5 & GP & 0:5:1:4 & 167, 205, 16 & 14.1 & 0.92 & 0.08 & 14.0--14.1 \\
18 & C6 & GP & 0:5:3:2 & 78, 170, 20 & 1.6 & 0.32 & 0.07 & 1.5--1.6 \\
19 & D1 & GP & 5:2:3:0 & 124, 53, 49.5 & 62.0 & 0.98 & 0.17 & 61.9--62.1 \\
20 & D2 & GP & 0:2:3:5 & 78, 177, 27.5 & 2.8 & 0.61 & 0.11 & 2.7--2.9 \\
21 & D3 & GP & 2:5:3:0 & 135.5, 77, 41 & 48.6 & 0.98 & 0.17 & 48.4--48.8 \\
22 & D4 & GP & 0:3:2:5 & 114, 194, 17 & 7.2 & 0.85 & 0.03 & 7.2--7.3 \\
23 & D5 & GP & 0:3:3:4 & 79, 177, 23.5 & 2.6 & 0.59 & 0.16 & 2.6--2.8 \\
24 & D6 & GP & 2:5:1:2 & 198, 97, 40 & 51.6 & 0.98 & 0.23 & 51.2--51.7 \\
\hline
\end{tabular*}
\end{table}

\FloatBarrier
\subsection*{B. Petroleum blue}
\begin{figure}[!htbp]
\centering
\includegraphics[width=0.95\textwidth]{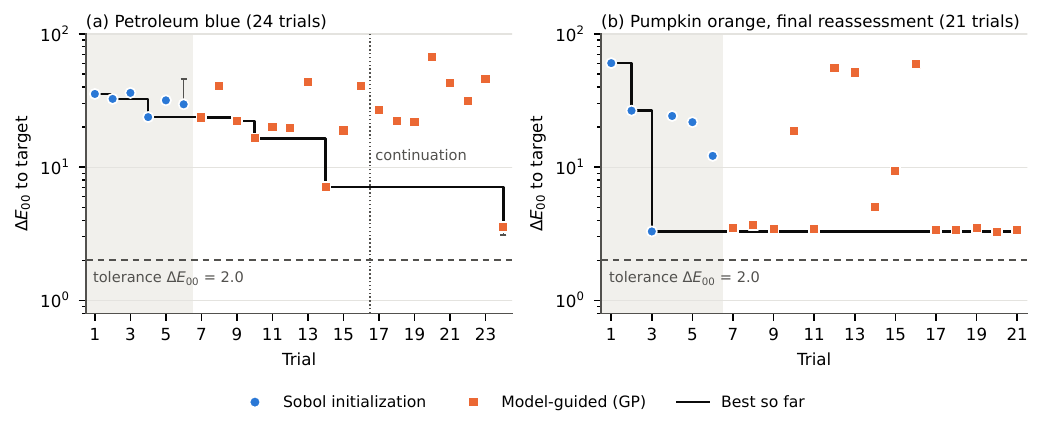}
\caption{Convergence of the petroleum blue and pumpkin orange campaigns. (a) Petroleum blue; the dotted line marks the start of the continuation in washed wells. (b) Pumpkin orange, plotted with the final reassessment values. Shaded regions mark Sobol initialization, error bars are bootstrap 2.5--97.5 percentile ranges, and the dashed line is the production tolerance of $\Delta E_{00} = 2.0$.}
\label{fig:blueorange}
\end{figure}

\begin{figure}[!htbp]
\centering
\includegraphics[width=0.62\textwidth]{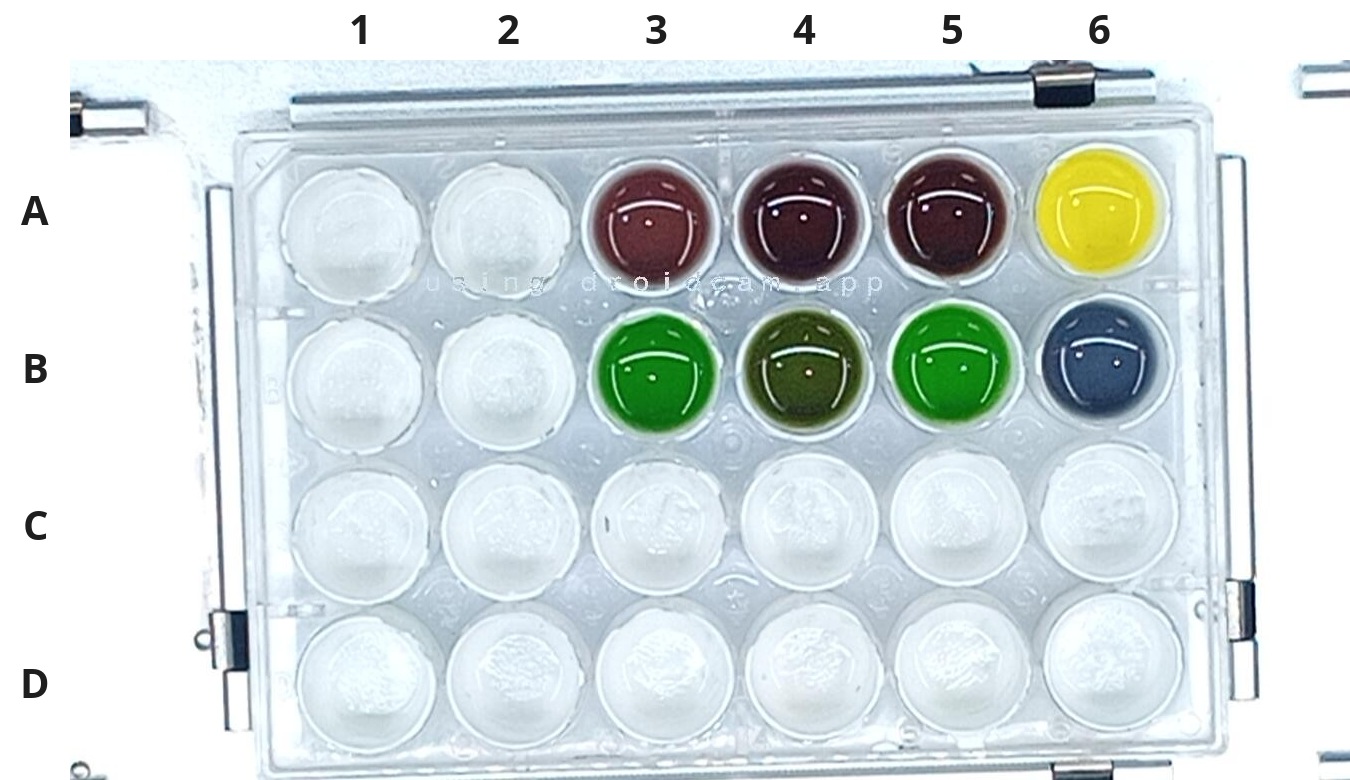}
\caption{Petroleum blue plate after the continuation, with trials 17--24 in the washed, reused wells A3--A6 and B3--B6. Rows A--D and columns 1--6 are labeled.}
\label{fig:plateblue}
\end{figure}

Following a changed phone position and deck layout, four preparations of 1:0:5:4 (100~\textmu L red, 500~\textmu L blue, 400~\textmu L water) defined the fixed target RGB [55, 78, 108] (Table~\ref{tbl:blueref}); all 24 accepted trials are listed in Table~\ref{tbl:bluetrials}, and the convergence and final plate are shown in Fig.~\ref{fig:blueorange}a and Fig.~\ref{fig:plateblue}. The six pairwise distances between reference wells ranged from 0.5 to 1.8, so the empirical allowance, max(1.1, largest pairwise distance), was 1.8, below the production tolerance of 2.0 selected for this campaign. The noise level supplied to the optimizer was 0.43. This four-well check did not demonstrate stability over the full production period.

The campaign completed 24 unique recipes: six Sobol, ten initial model-guided, and eight further model-guided trials. The first 16 trials used plate columns 3--6 with batch sizes of 4, 2, 4, 4, and 2. When that budget ended without reaching tolerance, wells A3--A6 and B3--B6 were washed for eight new preparations, and columns 1 and 2 remained excluded. All 16 completed observations and noise values were supplied to the continuation, which proposed four trials using 16 observations and then four using 20; no new Sobol trials were generated (Table~\ref{tbl:bluebatch}). The best Sobol score was 23.8, the best result of the initial model-guided phase was 7.1 at trial 14, and trial 24 reached 3.5, an 85.1\% reduction from initialization. Trial 24 evaluated 1:0:5:4, the reference formulation itself, yet measured RGB [64, 88, 116] against the target [55, 78, 108]; its estimated $L^*a^*b^*$ was [36.6, $-$1.0, $-$18.8] against [32.6, 0.3, $-$20.1]. The reference preparations were not training observations, so a formulation match did not guarantee the measured target color. Trial 6 showed acquisition instability, with a frame SD of 8.6 and per-frame distances from 28.8 to 48.9 around a median-RGB score of 29.6; the records do not identify whether camera behavior, sampling content, or illumination caused it. The 24 accepted preparations used 24~mL in total (red 5.5~mL, yellow 4.6~mL, blue 8.6~mL, water 5.3~mL) over 71 component transfers.

\begin{table}[!htbp]
\small
\caption{\ Petroleum blue reference preparations of 1:0:5:4; pooled target RGB [55, 78, 108]}
\label{tbl:blueref}
\begin{tabular*}{0.75\textwidth}{@{\extracolsep{\fill}}llll}
\hline
Well & Median camera RGB & $\Delta E_{00}$ to pooled target & Frame SD \\
\hline
C3 & 58, 77, 107 & 1.37 & 0.06 \\
C5 & 55, 77, 107 & 0.45 & 0.15 \\
D5 & 55, 79, 109 & 0.45 & 0.16 \\
A6 & 55, 79.5, 111 & 0.69 & 0.11 \\
\hline
\end{tabular*}
\end{table}

\begin{table}[!htbp]
\small
\caption{\ Petroleum blue batch structure. The last column gives the minimum within each batch, not the cumulative best}
\label{tbl:bluebatch}
\begin{tabular*}{0.75\textwidth}{@{\extracolsep{\fill}}llllll}
\hline
Batch & Trials & Count & Phase & Observations before & Batch minimum $\Delta E_{00}$ \\
\hline
1 & 1--4 & 4 & Sobol & 0 & 23.8 \\
2 & 5--6 & 2 & Sobol & 4 & 29.6 \\
3 & 7--10 & 4 & GP & 6 & 16.5 \\
4 & 11--14 & 4 & GP & 10 & 7.1 \\
5 & 15--16 & 2 & GP & 14 & 18.8 \\
6 & 17--20 & 4 & GP (continuation) & 16 & 21.9 \\
7 & 21--24 & 4 & GP (continuation) & 20 & 3.5 \\
\hline
\end{tabular*}
\end{table}

\begin{table}[!htbp]
\footnotesize
\caption{\ Petroleum blue: all 24 accepted trials (target RGB [55, 78, 108], $T = 2.0$, noise level supplied to the optimizer 0.43). Trials 17--24 are fresh preparations in washed, reused wells}
\label{tbl:bluetrials}
\begin{tabular*}{\textwidth}{@{\extracolsep{\fill}}lllllllll}
\hline
Trial \# & Well & Phase & R:Y:B:W & Median camera RGB & $\Delta E_{00}$ & $D$ & Frame SD & Bootstrap 2.5--97.5\% \\
\hline
1 & A3 & Sobol & 3:4:1:2 & 146, 37, 37 & 35.4 & 0.94 & 0.05 & 35.4--35.5 \\
2 & A4 & Sobol & 2:2:2:4 & 119, 46, 36 & 32.5 & 0.94 & 0.07 & 32.4--32.5 \\
3 & A5 & Sobol & 1:4:3:2 & 92, 83, 35 & 36.1 & 0.95 & 0.22 & 36.1--36.4 \\
4 & A6 & Sobol & 4:0:4:2 & 78.5, 47, 58 & 23.8 & 0.92 & 0.16 & 23.5--23.9 \\
5 & B3 & Sobol & 4:2:2:2 & 123, 43, 49 & 31.7 & 0.94 & 0.07 & 31.6--31.8 \\
6 & B4 & Sobol & 3:2:3:2 & 106, 45, 49 & 29.6 & 0.93 & 8.63 & 29.3--46.1 \\
7 & B5 & GP & 0:0:5:5 & 33, 137, 206 & 23.6 & 0.92 & 0.14 & 23.6--23.8 \\
8 & B6 & GP & 5:0:0:5 & 215, 54, 67 & 40.8 & 0.95 & 0.10 & 40.8--41.0 \\
9 & C3 & GP & 5:0:5:0 & 73, 47, 59 & 22.3 & 0.91 & 0.14 & 22.0--22.3 \\
10 & C4 & GP & 2:0:5:3 & 77, 62, 80 & 16.5 & 0.88 & 0.30 & 16.5--16.6 \\
11 & C5 & GP & 2:3:5:0 & 81, 65, 62 & 19.9 & 0.90 & 0.23 & 19.6--20.1 \\
12 & C6 & GP & 3:2:5:0 & 81, 61, 68 & 19.7 & 0.90 & 0.22 & 19.7--19.9 \\
13 & D3 & GP & 0:2:5:3 & 29, 139, 49 & 43.5 & 0.95 & 0.07 & 43.4--43.5 \\
14 & D4 & GP & 1:0:4:5 & 80, 96, 118 & 7.1 & 0.72 & 0.11 & 7.1--7.3 \\
15 & D5 & GP & 3:0:5:2 & 78, 59, 71 & 18.8 & 0.89 & 0.26 & 18.7--19.2 \\
16 & D6 & GP & 5:5:0:0 & 216, 62.5, 75.5 & 40.7 & 0.95 & 0.39 & 40.7--41.0 \\
17 & A3 & GP & 2:0:3:5 & 85, 43, 48 & 27.0 & 0.93 & 0.10 & 27.0--27.1 \\
18 & A4 & GP & 4:0:5:1 & 54, 36, 44 & 22.2 & 0.91 & 0.07 & 22.2--22.4 \\
19 & A5 & GP & 4:1:5:0 & 63, 42, 51 & 21.9 & 0.91 & 0.15 & 21.7--22.1 \\
20 & A6 & GP & 0:5:0:5 & 216, 206, 40 & 66.9 & 0.97 & 0.06 & 66.9--67.0 \\
21 & B3 & GP & 0:5:5:0 & 27, 125, 32 & 42.5 & 0.95 & 0.07 & 42.5--42.6 \\
22 & B4 & GP & 1:4:5:0 & 63, 82, 45 & 31.6 & 0.94 & 0.30 & 31.1--31.6 \\
23 & B5 & GP & 0:5:4:1 & 28, 140.5, 31 & 45.7 & 0.96 & 0.12 & 45.6--45.8 \\
24 & B6 & GP & 1:0:5:4 & 64, 88, 116 & 3.5 & 0.44 & 0.23 & 3.1--3.6 \\
\hline
\end{tabular*}
\end{table}

\FloatBarrier
\subsection*{C. Pumpkin orange}
\begin{figure}[!htbp]
\centering
\includegraphics[width=0.62\textwidth]{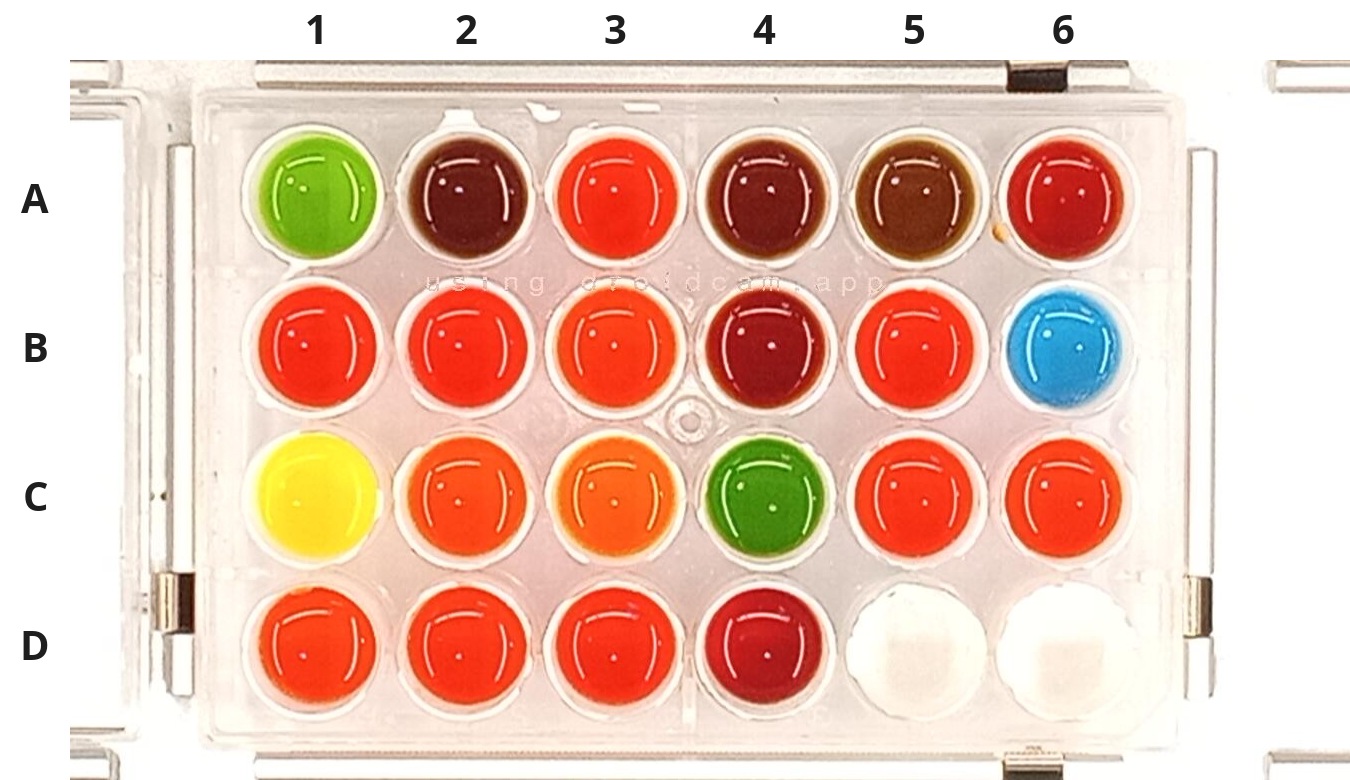}
\caption{Pumpkin orange plate at the final reassessment. Wells A1--D3 hold the 21 accepted trials; D4 held an extra preparation and was excluded. Rows A--D and columns 1--6 are labeled.}
\label{fig:plateorange}
\end{figure}

Five preparations of 2:5:0:3 (200~\textmu L red, 500~\textmu L yellow, 300~\textmu L water) defined the pooled target RGB [230, 70, 31] (Table~\ref{tbl:orangeref}); the convergence on the final values and the plate at reassessment are shown in Fig.~\ref{fig:blueorange}b and Fig.~\ref{fig:plateorange}. The ten pairwise distances between reference wells ranged from 0.8 to 4.7, so the empirical allowance of 4.7 exceeded the production tolerance of 2.0 selected for this campaign; the criterion therefore did not represent demonstrated two-unit repeatability. The noise level supplied to the optimizer was 0.99.

The original plan was six Sobol and 18 model-guided trials in batches of three. The operator closed the campaign after 21 accepted recipes (six Sobol, 15 model-guided), giving a recipe coverage of 14.4\%; well D4 contained an extra preparation during reassessment and was excluded. A camera interruption at well B1 produced a superseded score of 35.0. Recovery retained the six completed Sobol observations, regenerated and verified the original first model-guided batch, and measured the retained B1 and B2 mixtures without dispensing again; pipetting resumed at B3.

\begin{figure}[!htbp]
\centering
\includegraphics[width=0.8\textwidth]{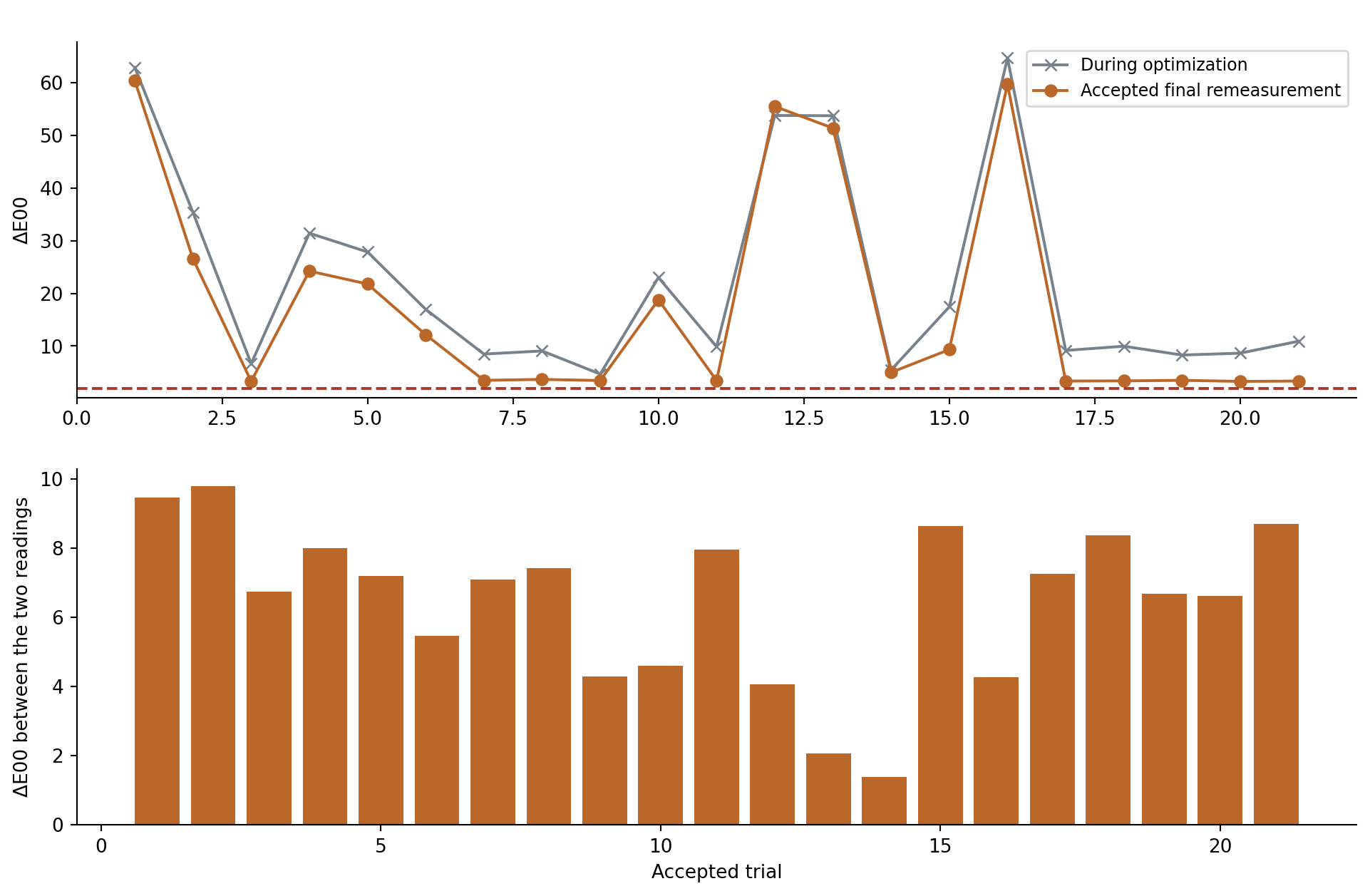}
\caption{Pumpkin orange measured on two occasions. Top: target distance of each accepted trial during optimization and in the final reassessment, which the optimizer did not see. Bottom: $\Delta E_{00}$ between the two readings of the same retained mixture.}
\label{fig:orangeoccasions}
\end{figure}

The optimizer used online readings, and its best accepted online result was trial 9 in B3 at $\Delta E_{00} = 4.7$. After production stopped, ten new full-plate frames remeasured A1--D4, and the operator accepted A1--D3 as final results (Table~\ref{tbl:orangetrials}). These 210 per-well readings share ten acquisition times and were not fed back to the optimizer or used to alter historical stopping decisions. On the final values, trial 20 in D2 ranked best (4:3:0:3, RGB [248, 62, 32], $\Delta E_{00} = 3.3$, $D = 0.39$), but the best Sobol result, trial 3 at 3.3, was less than 0.01 higher, so this retrospective comparison does not establish an improvement of the best point over initialization. The changes between the two measurement occasions were much larger than the ten-frame variation (Table~\ref{tbl:orangeoccasions}, Fig.~\ref{fig:orangeoccasions}), and neither the records nor the images separate camera settings, illumination, geometry, settling, and physical changes in the mixtures as causes. The accepted preparations used 21~mL (red 6.2~mL, yellow 6.3~mL, blue 2.8~mL, water 5.7~mL) over 61 component transfers.

\begin{table}[!htbp]
\small
\caption{\ Pumpkin orange reference preparations of 2:5:0:3; pooled target RGB [230, 70, 31]}
\label{tbl:orangeref}
\begin{tabular*}{0.75\textwidth}{@{\extracolsep{\fill}}llll}
\hline
Well & Median camera RGB & $\Delta E_{00}$ to pooled target & Frame SD \\
\hline
A1 & 225, 62, 25 & 1.99 & 0.05 \\
A6 & 230, 70, 32 & 0.18 & 0.15 \\
B3 & 236.5, 69, 26 & 1.34 & 0.11 \\
C4 & 230, 73, 31 & 0.69 & 0.08 \\
D6 & 235, 82, 40 & 2.67 & 0.02 \\
\hline
\end{tabular*}
\end{table}

\begin{table}[!htbp]
\small
\caption{\ Pumpkin orange: comparison of the online measurements used by the optimizer with the final reassessment}
\label{tbl:orangeoccasions}
\begin{tabular*}{0.75\textwidth}{@{\extracolsep{\fill}}ll}
\hline
Measurement feature & Recorded result \\
\hline
Maximum change in target distance between occasions & 8.8 $\Delta E_{00}$ \\
Trial 20, online RGB and target distance & [252, 68, 66], 8.6 \\
Trial 20, final RGB and target distance & [248, 62, 32], 3.3 \\
Difference between the two trial 20 colors & 6.6 $\Delta E_{00}$ \\
Final results below $\Delta E_{00} = 5$ & GP 10 of 15; Sobol 1 of 6 \\
\hline
\end{tabular*}
\end{table}

\begin{table}[!htbp]
\footnotesize
\caption{\ Pumpkin orange: all 21 accepted trials, final reassessment values (target RGB [230, 70, 31], $T = 2.0$, noise level supplied to the optimizer 0.99)}
\label{tbl:orangetrials}
\begin{tabular*}{\textwidth}{@{\extracolsep{\fill}}lllllllll}
\hline
Trial \# & Well & Phase & R:Y:B:W & Median camera RGB & $\Delta E_{00}$ & $D$ & Frame SD & Bootstrap 2.5--97.5\% \\
\hline
1 & A1 & Sobol & 0:3:2:5 & 125, 181, 25 & 60.4 & 0.97 & 0.10 & 60.3--60.4 \\
2 & A2 & Sobol & 3:1:5:1 & 112, 41, 34 & 26.5 & 0.93 & 0.10 & 26.3--26.5 \\
3 & A3 & Sobol & 4:4:0:2 & 247, 56.5, 31 & 3.3 & 0.39 & 0.06 & 3.3--3.3 \\
4 & A4 & Sobol & 3:2:4:1 & 125, 39, 32 & 24.2 & 0.92 & 0.12 & 24.1--24.3 \\
5 & A5 & Sobol & 2:4:4:0 & 127, 55, 30 & 21.8 & 0.91 & 0.10 & 21.7--21.8 \\
6 & A6 & Sobol & 4:1:1:4 & 192, 39, 34 & 12.2 & 0.84 & 0.14 & 12.2--12.4 \\
7 & B1 & GP & 5:5:0:0 & 247, 55, 32 & 3.5 & 0.43 & 0.08 & 3.4--3.6 \\
8 & B2 & GP & 5:0:0:5 & 248, 54, 32 & 3.7 & 0.45 & 0.05 & 3.6--3.7 \\
9 & B3 & GP & 3:5:0:2 & 249, 67, 31 & 3.5 & 0.42 & 0.01 & 3.5--3.5 \\
10 & B4 & GP & 5:3:2:0 & 159, 37, 36 & 18.8 & 0.89 & 0.10 & 18.8--18.9 \\
11 & B5 & GP & 5:3:0:2 & 246, 57, 34 & 3.4 & 0.42 & 0.06 & 3.4--3.5 \\
12 & B6 & GP & 0:0:5:5 & 52, 164, 213 & 55.5 & 0.96 & 0.07 & 55.4--55.5 \\
13 & C1 & GP & 0:5:0:5 & 255, 230, 47 & 51.4 & 0.96 & 0.00 & 51.4--51.4 \\
14 & C2 & GP & 2:5:0:3 & 251, 80, 32 & 5.0 & 0.60 & 0.07 & 5.0--5.1 \\
15 & C3 & GP & 1:5:0:4 & 250, 102, 32 & 9.3 & 0.79 & 0.04 & 9.3--9.4 \\
16 & C4 & GP & 0:5:5:0 & 83, 146, 24 & 59.7 & 0.97 & 0.15 & 59.7--59.9 \\
17 & C5 & GP & 4:1:0:5 & 248, 62, 33 & 3.4 & 0.40 & 0.03 & 3.3--3.4 \\
18 & C6 & GP & 4:5:0:1 & 248, 67, 34 & 3.4 & 0.41 & 0.01 & 3.4--3.4 \\
19 & D1 & GP & 3:2:0:5 & 249, 68, 31 & 3.5 & 0.43 & 0.06 & 3.5--3.5 \\
20 & D2 & GP & 4:3:0:3 & 248, 62, 32 & 3.3 & 0.39 & 0.02 & 3.3--3.3 \\
21 & D3 & GP & 5:1:0:4 & 246, 56, 32.5 & 3.4 & 0.40 & 0.06 & 3.3--3.4 \\
\hline
\end{tabular*}
\end{table}

\FloatBarrier
\subsection*{D. Initial camera run on RAINBOT\textsuperscript{TM}}
\begin{table}[!htbp]
\small
\caption{\ Initial camera run on RAINBOT\textsuperscript{TM} against a green target (RGB [28, 154, 17]). Recipes are R:Y:B:W transfer units (each component 0--5, sum $\leq$ 10); $\Delta E_{00}$ is the CIEDE2000 difference from the target}
\label{tbl:run2}
\begin{tabular*}{0.75\textwidth}{@{\extracolsep{\fill}}llllllll}
\hline
Trial \# & Phase & R:Y:B:W & $\Delta E_{00}$ & Trial \# & Phase & R:Y:B:W & $\Delta E_{00}$ \\
\hline
1 & Sobol & 2:5:0:3 & 69.4 & 13 & GP & 1:0:3:5 & 40.3 \\
2 & Sobol & 1:2:4:1 & 32.0 & 14 & GP & 3:0:5:2 & 51.0 \\
3 & Sobol & 0:4:5:0 & 2.4 & 15 & GP & 3:2:2:1 & 55.2 \\
4 & Sobol & 2:1:1:4 & 63.0 & 16 & GP & 0:4:4:2 & 28.2 \\
5 & Sobol & 4:2:1:0 & 70.3 & 17 & GP & 2:2:2:0 & 46.2 \\
6 & Sobol & 0:0:1:1 & 43.6 & 18 & GP & 2:1:5:2 & 39.8 \\
7 & GP & 0:5:5:0 & 6.3 & 19 & GP & 0:2:4:1 & 30.4 \\
8 & GP & 0:3:5:0 & 9.6 & 20 & GP & 2:0:4:2 & 46.8 \\
9 & GP & 0:4:5:1 & 18.5 & 21 & GP & 1:1:3:4 & 35.8 \\
10 & GP & 0:4:3:0 & 24.7 & 22 & GP & 3:4:0:1 & 36.6 \\
11 & GP & 0:4:3:2 & 26.3 & 23 & GP & 1:4:3:0 & 35.1 \\
12 & GP & 1:4:5:0 & 23.3 & 24 & GP & 4:1:4:0 & 52.7 \\
\hline
\end{tabular*}
\end{table}

\subsection*{E. Gravimetric raw data}
\begin{table*}[!htbp]
\footnotesize
\caption{\ Individual dispensed volumes for the gravimetric validation of RAINBOT\textsuperscript{TM}, in \textmu L: evaporation-corrected value with the uncorrected value in parentheses, in run order within each day. On day 2, four first attempts were repeated and replaced: 100~\textmu L (no dispense; 0.0), 100~\textmu L (short; 90.6), 200~\textmu L (wet washed tip; 217.1), and 400~\textmu L (wet washed tip; 437.0)}
\label{tbl:rawbalance}
\begin{tabular*}{\textwidth}{@{\extracolsep{\fill}}llll}
\hline
Target / \textmu L & Day 1 & Day 2 & Day 3 \\
\hline
100 & 98.6 (93.0), 91.9 (79.7), 93.9 (88.3) & 106.8 (100.5), 101.5 (96.5), 104.1 (93.7) & 121.7 (115.7), 103.0 (98.4), 105.4 (100.8), 105.6 (101.0) \\
200 & 202.4 (197.4), 183.2 (178.7), 189.4 (184.2) & 208.6 (199.4), 202.3 (195.7), 199.1 (190.7) & 208.7 (204.7), 209.2 (204.7), 198.8 (193.4) \\
300 & 292.7 (283.4), 305.6 (299.4), 293.2 (288.1) & 314.7 (305.1), 308.3 (302.8), 300.2 (292.5) & 307.9 (303.6), 331.6 (322.2), 295.7 (291.5) \\
400 & 392.0 (386.0), 395.2 (390.6), 389.7 (385.2) & 401.4 (382.5), 399.4 (392.2), 396.7 (389.1) & 411.2 (407.1), 413.1 (408.2), 398.2 (388.8) \\
500 & 492.6 (485.4), 488.0 (481.9), 483.0 (476.3) & 499.3 (493.8), 495.0 (489.7), 498.4 (491.2) & 516.0 (510.1), 510.8 (505.3), 504.4 (499.9) \\
700 & 695.1 (687.3), 699.7 (695.5), 686.5 (678.0) & 701.4 (695.2), 701.9 (694.8), 701.6 (695.5) & 711.4 (705.5), 715.1 (710.5), 709.2 (703.5) \\
1000 & 974.4 (959.3), 975.9 (965.0), 978.3 (972.1) & 999.5 (984.2), 991.5 (984.9), 987.3 (982.8) & 1005.2 (990.4), 992.8 (988.8), 986.6 (982.0) \\
\hline
\end{tabular*}
\end{table*}

\FloatBarrier
\subsection*{F. Emerald green ten-well measurement check}
\begin{figure}[!htbp]
\centering
\includegraphics[width=0.62\textwidth]{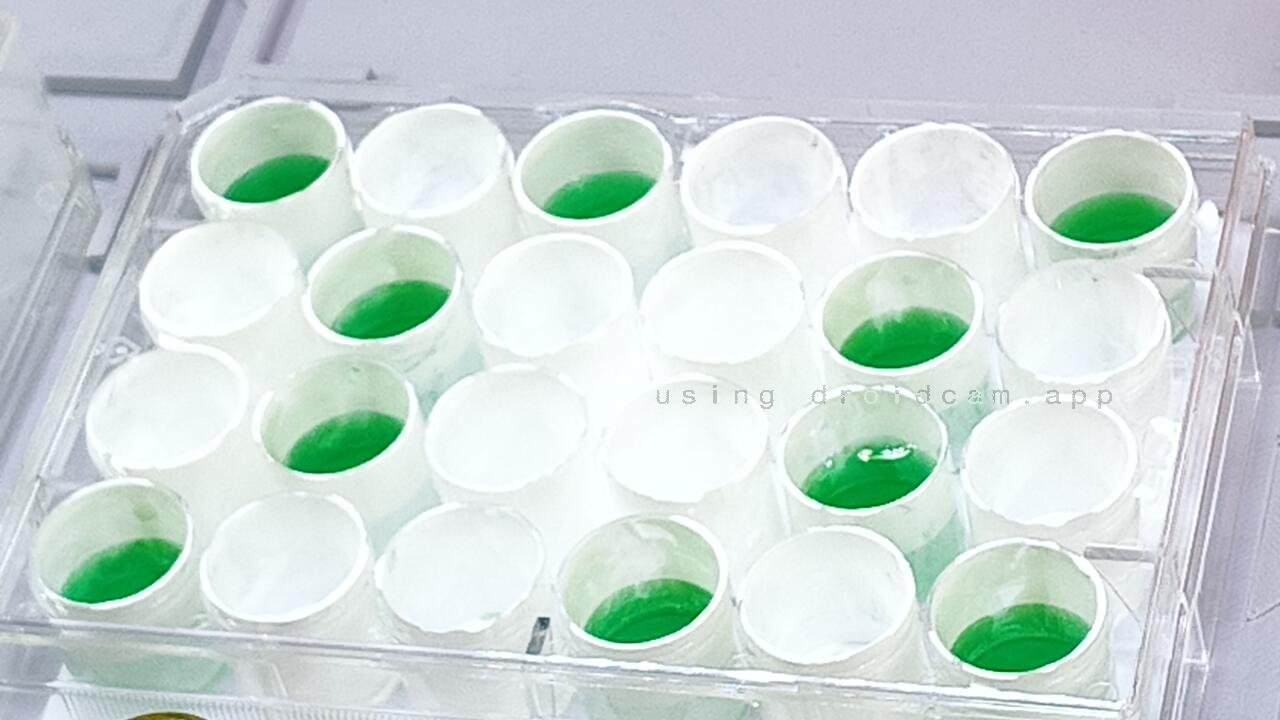}
\caption{Emerald green plate photographed at the end of the ten-well check. The ten replicate wells (A1, A3, A6, B2, B5, C2, C5, D1, D4, and D6) hold 0:4:5:1, and the other wells hold water. The photograph retains acquisition reflections and the DroidCam watermark.}
\label{fig:plateemerald}
\end{figure}

This study was run on RAINBOT\textsuperscript{TM} with the mobile phone camera (DroidCam stream) on 15 November 2026. The emerald green recipe 0:4:5:1 (400~\textmu L yellow, 500~\textmu L blue, 100~\textmu L water) was prepared independently in ten wells, and the remaining 14 wells received 1000~\textmu L of water. Well A1 was designated as the reference well, and each well was read ten times at target verification, immediately before the run. The pooled target, RGB [8.2, 170.3, 40.4] (\#08AA28; $L^* = 60.9$, $a^* = -61.5$, $b^* = 52.9$), is the componentwise median of the ten replicate wells. Table~\ref{tbl:emeraldref} lists the replicate wells, Table~\ref{tbl:emeraldpair} the full pairwise matrix, Table~\ref{tbl:emeraldtrials} all 24 accepted readings, Table~\ref{tbl:emeraldbatch} the batch structure, and Table~\ref{tbl:emeraldclosest} the replicate closest to the pooled target; Fig.~\ref{fig:plateemerald} shows the plate at completion. The operating threshold is $T = 5.5$, and $D$ is recomputed with eqn~(\ref{eq:distance}) for this threshold. Bootstrap ranges use 30\,000 whole-vector resamples of the ten RGB readings per well. The GP observation-noise proxy for this run was 0.96. All 24 wells used 24~mL in total (yellow 4.0~mL, blue 5.0~mL, water 15.0~mL; no red), and the 24 wells cover 2 of the 146 feasible recipes.

\begin{table}[!htbp]
\small
\caption{\ Emerald green replicate wells of 0:4:5:1 (reference well A1). Distances are $\Delta E_{00}$ to the reference well and to the pooled target RGB [8.2, 170.3, 40.4]}
\label{tbl:emeraldref}
\begin{tabular*}{0.9\textwidth}{@{\extracolsep{\fill}}lllllll}
\hline
Well & Role & Median camera RGB & HEX & $\Delta E_{00}$ to A1 & $\Delta E_{00}$ to pooled target & Frame SD \\
\hline
A1 & Reference & 8.2, 170.2, 45.8 & \#08AA2E & 0.00 & 0.56 & 0.15 \\
A3 & Replicate & 10.2, 170.4, 44.6 & \#0AAA2D & 0.17 & 0.40 & 0.05 \\
A6 & Replicate & 10.2, 182.8, 52.3 & \#0AB734 & 3.62 & 3.62 & 0.09 \\
B2 & Replicate & 6.5, 168.5, 33.3 & \#06A821 & 1.22 & 0.78 & 0.07 \\
B5 & Replicate & 8.1, 170.1, 35.3 & \#08AA23 & 1.03 & 0.47 & 0.06 \\
C2 & Replicate & 7.0, 169.0, 33.8 & \#07A922 & 1.15 & 0.66 & 0.04 \\
C5 & Replicate & 7.9, 174.3, 34.9 & \#08AE23 & 1.77 & 1.39 & 0.07 \\
D1 & Replicate & 7.0, 165.4, 39.6 & \#07A528 & 1.49 & 1.50 & 0.06 \\
D4 & Replicate & 8.4, 172.1, 44.3 & \#08AC2C & 0.61 & 0.59 & 0.08 \\
D6 & Replicate & 8.9, 173.7, 41.3 & \#09AE29 & 1.23 & 1.00 & 0.07 \\
\hline
\end{tabular*}
\end{table}

\begin{table}[!htbp]
\small
\caption{\ Emerald green: symmetric pairwise $\Delta E_{00}$ matrix between the ten replicate wells (reference well A1). The last column is the mean distance of each well to the other nine}
\label{tbl:emeraldpair}
\begin{tabular*}{0.9\textwidth}{@{\extracolsep{\fill}}lrrrrrrrrrrr}
\hline
Well & A1 & A3 & A6 & B2 & B5 & C2 & C5 & D1 & D4 & D6 & Mean \\
\hline
A1 (ref.) & 0.00 & 0.17 & 3.62 & 1.22 & 1.03 & 1.15 & 1.77 & 1.49 & 0.61 & 1.23 & 1.37 \\
A3 & 0.17 & 0.00 & 3.56 & 1.12 & 0.87 & 1.03 & 1.61 & 1.53 & 0.49 & 1.09 & 1.27 \\
A6 & 3.62 & 3.56 & 0.00 & 4.27 & 3.78 & 4.11 & 2.68 & 5.08 & 3.08 & 2.66 & 3.65 \\
B2 & 1.22 & 1.12 & 4.27 & 0.00 & 0.49 & 0.17 & 1.75 & 1.19 & 1.36 & 1.61 & 1.46 \\
B5 & 1.03 & 0.87 & 3.78 & 0.49 & 0.00 & 0.32 & 1.29 & 1.56 & 0.96 & 1.13 & 1.27 \\
C2 & 1.15 & 1.03 & 4.11 & 0.17 & 0.32 & 0.00 & 1.59 & 1.32 & 1.22 & 1.45 & 1.37 \\
C5 & 1.77 & 1.61 & 2.68 & 1.75 & 1.29 & 1.59 & 0.00 & 2.83 & 1.21 & 0.64 & 1.71 \\
D1 & 1.49 & 1.53 & 5.08 & 1.19 & 1.56 & 1.32 & 2.83 & 0.00 & 2.00 & 2.50 & 2.17 \\
D4 & 0.61 & 0.49 & 3.08 & 1.36 & 0.96 & 1.22 & 1.21 & 2.00 & 0.00 & 0.63 & 1.28 \\
D6 & 1.23 & 1.09 & 2.66 & 1.61 & 1.13 & 1.45 & 0.64 & 2.50 & 0.63 & 0.00 & 1.44 \\
\hline
\end{tabular*}
\end{table}

\begin{table}[!htbp]
\footnotesize
\caption{\ Emerald green: all 24 accepted readings (pooled target RGB [8.2, 170.3, 40.4], $T = 5.5$). Recipe 0:4:5:1 is the emerald green replicate and 0:0:0:10 is water}
\label{tbl:emeraldtrials}
\begin{tabular*}{\textwidth}{@{\extracolsep{\fill}}llllllll}
\hline
Trial \# & Well & R:Y:B:W & Median camera RGB & $\Delta E_{00}$ & $D$ & Frame SD & Bootstrap 2.5--97.5\% \\
\hline
1 & A1 & 0:4:5:1 & 8.2, 170.2, 45.8 & 0.6 & 0.00 & 0.15 & 0.4--0.7 \\
2 & A2 & 0:0:0:10 & 227.5, 226.0, 213.9 & 33.7 & 0.84 & 0.04 & 33.7--33.8 \\
3 & A3 & 0:4:5:1 & 10.2, 170.4, 44.6 & 0.4 & 0.00 & 0.05 & 0.4--0.4 \\
4 & A4 & 0:0:0:10 & 224.3, 225.3, 217.3 & 33.9 & 0.84 & 0.03 & 33.9--33.9 \\
5 & A5 & 0:0:0:10 & 228.4, 229.4, 221.4 & 34.4 & 0.84 & 0.03 & 34.4--34.5 \\
6 & A6 & 0:4:5:1 & 10.2, 182.8, 52.3 & 3.6 & 0.00 & 0.09 & 3.6--3.7 \\
7 & B1 & 0:0:0:10 & 221.7, 225.2, 216.2 & 33.3 & 0.84 & 0.09 & 33.3--33.4 \\
8 & B2 & 0:4:5:1 & 6.5, 168.5, 33.3 & 0.8 & 0.00 & 0.07 & 0.7--0.8 \\
9 & B3 & 0:0:0:10 & 229.2, 230.2, 222.2 & 34.5 & 0.84 & 0.03 & 34.5--34.6 \\
10 & B4 & 0:0:0:10 & 221.3, 225.5, 216.2 & 33.2 & 0.83 & 0.09 & 33.2--33.3 \\
11 & B5 & 0:4:5:1 & 8.1, 170.1, 35.3 & 0.5 & 0.00 & 0.06 & 0.4--0.5 \\
12 & B6 & 0:0:0:10 & 231.2, 232.2, 224.2 & 34.8 & 0.84 & 0.02 & 34.8--34.8 \\
13 & C1 & 0:0:0:10 & 223.8, 225.9, 217.5 & 33.7 & 0.84 & 0.08 & 33.6--33.8 \\
14 & C2 & 0:4:5:1 & 7.0, 169.0, 33.8 & 0.7 & 0.00 & 0.04 & 0.6--0.7 \\
15 & C3 & 0:0:0:10 & 229.7, 230.7, 222.7 & 34.6 & 0.84 & 0.02 & 34.6--34.6 \\
16 & C4 & 0:0:0:10 & 224.5, 226.4, 218.0 & 33.8 & 0.84 & 0.11 & 33.7--33.9 \\
17 & C5 & 0:4:5:1 & 7.9, 174.3, 34.9 & 1.4 & 0.00 & 0.07 & 1.3--1.4 \\
18 & C6 & 0:0:0:10 & 237.7, 238.7, 230.7 & 35.7 & 0.85 & 0.02 & 35.7--35.7 \\
19 & D1 & 0:4:5:1 & 7.0, 165.4, 39.6 & 1.5 & 0.00 & 0.06 & 1.4--1.5 \\
20 & D2 & 0:0:0:10 & 224.8, 225.7, 214.9 & 33.5 & 0.84 & 0.09 & 33.4--33.6 \\
21 & D3 & 0:0:0:10 & 221.8, 222.8, 214.8 & 33.5 & 0.84 & 0.02 & 33.5--33.6 \\
22 & D4 & 0:4:5:1 & 8.4, 172.1, 44.3 & 0.6 & 0.00 & 0.08 & 0.5--0.7 \\
23 & D5 & 0:0:0:10 & 226.6, 227.6, 219.5 & 34.2 & 0.84 & 0.03 & 34.2--34.2 \\
24 & D6 & 0:4:5:1 & 8.9, 173.7, 41.3 & 1.0 & 0.00 & 0.07 & 1.0--1.1 \\
\hline
\end{tabular*}
\end{table}

\begin{table}[!htbp]
\small
\caption{\ Emerald green batch structure. Batch minimum is the smallest $\Delta E_{00}$ within the batch; running best is cumulative}
\label{tbl:emeraldbatch}
\begin{tabular*}{0.75\textwidth}{@{\extracolsep{\fill}}llllll}
\hline
Batch & Trials & Count & Readings before & Batch minimum $\Delta E_{00}$ & Running best $\Delta E_{00}$ \\
\hline
1 & 1--4 & 4 & 0 & 0.4 & 0.4 \\
2 & 5--8 & 4 & 4 & 0.8 & 0.4 \\
3 & 9--12 & 4 & 8 & 0.5 & 0.4 \\
4 & 13--16 & 4 & 12 & 0.7 & 0.4 \\
5 & 17--20 & 4 & 16 & 1.4 & 0.4 \\
6 & 21--24 & 4 & 20 & 0.6 & 0.4 \\
\hline
\end{tabular*}
\end{table}

\begin{table}[!htbp]
\small
\caption{\ Emerald green: pooled target compared with the closest replicate (trial 3, well A3)}
\label{tbl:emeraldclosest}
\begin{tabular*}{0.75\textwidth}{@{\extracolsep{\fill}}llll}
\hline
Quantity & Pooled target & Trial 3, A3 & Difference \\
\hline
R:Y:B:W & 0:4:5:1 & 0:4:5:1 & --- \\
RGB & 8.2, 170.3, 40.4 & 10.2, 170.4, 44.6 & +2.0, +0.1, +4.2 \\
$L^*$ & 60.9 & 60.9 & +0.1 \\
$a^*$ & $-$61.5 & $-$61.1 & +0.4 \\
$b^*$ & 52.9 & 51.5 & $-$1.4 \\
Chroma $C^*$ & 81.1 & 79.9 & $-$1.2 \\
Hue angle / $^\circ$ & 139.3 & 139.9 & +0.6 \\
$\Delta E_{00}$ & 0 & 0.4 & 0.4 \\
\hline
\end{tabular*}
\end{table}

\FloatBarrier
\subsection*{G. Multi-color ten-well camera repeatability}
Table~\ref{tbl:multicolor} summarizes the 16 runs of the multi-color check (22 and 23 November 2026). Each color was a single hand-prepared batch dispensed into wells A1, A3, A6, B2, B5, C2, C5, D1, D4, and D6, with A1 as the reference. Distances compare per-well median colors by CIEDE2000.

\begin{table*}[!htbp]
\footnotesize
\caption{\ Multi-color ten-well camera repeatability. Mean pair is the mean $\pm$ SD of the 45 pairwise $\Delta E_{00}$; the color passes overall if it is below $T = 5.5$. Wells failing are those at or above 5.5 from reference A1}
\label{tbl:multicolor}
\begin{tabular*}{\textwidth}{@{\extracolsep{\fill}}lllllllll}
\hline
Color & R:Y:B:W & Session & Lights & Mean pair & Max pair & Max vs A1 & Pairs $< T$ & Wells failing vs A1 \\
\hline
Grass green & 0:5:2:3 & Evening & Off & $2.80 \pm 1.63$ & 7.49 & 4.52 & 42/45 & None \\
Grass green & 0:5:2:3 & Evening & On & $3.11 \pm 2.04$ & 7.88 & 4.44 & 38/45 & None \\
Petroleum blue & 1:0:5:4 & Evening & Off & $3.32 \pm 1.77$ & 7.70 & 4.68 & 39/45 & None \\
Petroleum blue & 1:0:5:4 & Evening & On & $2.90 \pm 1.81$ & 7.63 & 5.67 & 40/45 & B5 \\
Pumpkin orange & 2:5:0:3 & Evening & Off & $2.89 \pm 1.63$ & 7.48 & 3.97 & 42/45 & None \\
Pumpkin orange & 2:5:0:3 & Evening & On & $2.99 \pm 1.71$ & 7.30 & 4.71 & 41/45 & None \\
Orange-red & 3:5:0:2 & Evening & Off & $3.54 \pm 1.97$ & 6.47 & 6.09 & 35/45 & B2, C5, D6 \\
Orange-red & 3:5:0:2 & Evening & On & $3.60 \pm 2.18$ & 7.43 & 7.43 & 34/45 & B2, B5, C5, D6 \\
Emerald green & 0:4:5:1 & Evening & Off & $3.38 \pm 1.94$ & 7.72 & 7.72 & 36/45 & B2, B5, C5, D4, D6 \\
Emerald green & 0:4:5:1 & Evening & On & $2.86 \pm 1.56$ & 6.43 & 6.43 & 41/45 & B2, B5, D6 \\
Grass green & 0:5:2:3 & Daylight & Off & $3.10 \pm 2.54$ & 9.56 & 7.45 & 37/45 & A6 \\
Grass green & 0:5:2:3 & Daylight & On & $2.14 \pm 1.33$ & 5.50 & 3.13 & 44/45 & None \\
Petroleum blue & 1:0:5:4 & Daylight & Off & $4.96 \pm 3.05$ & 13.30 & 10.33 & 27/45 & B5, D4, D6 \\
Petroleum blue & 1:0:5:4 & Daylight & On & $4.40 \pm 2.82$ & 12.01 & 12.01 & 32/45 & B2, B5, C2, C5, D1, D4, D6 \\
Pumpkin orange & 2:5:0:3 & Daylight & Off & $4.20 \pm 2.72$ & 9.60 & 9.60 & 34/45 & A3, B5, C2, C5, D4, D6 \\
Pumpkin orange & 2:5:0:3 & Daylight & On & $4.51 \pm 3.42$ & 13.74 & 13.74 & 31/45 & A3, A6, B2, B5, C2, C5, D1, D4, D6 \\
\hline
\end{tabular*}
\end{table*}

\end{document}